\documentclass{article}

\usepackage{microtype}
\usepackage{graphicx}
\usepackage{subfigure}

\usepackage{booktabs} % for professional tables

\usepackage{algorithm}
\usepackage{amsfonts} % 提供\mathbb命令
\usepackage{amssymb}  % 提供更多的数学符号
\usepackage{enumitem}

\usepackage{amsmath}  % 提供数学环境和命令
\usepackage{amssymb}  % 提供额外的数学符号
\usepackage{bm}       % 提供粗体斜体命令 \bm
\usepackage{amsmath}
\usepackage{xcolor}
\usepackage{hyperref}

\usepackage[accepted]{icml2024}

\usepackage{amsmath}
\usepackage{amssymb}
\usepackage{mathtools}
\usepackage{amsthm}
\usepackage{multirow}
\usepackage{ulem}
\usepackage[capitalize,noabbrev]{cleveref}

\theoremstyle{plain}

\theoremstyle{definition}

\theoremstyle{remark}

\DeclareMathOperator*{\argmax}{arg\,max}

\usepackage[textsize=tiny]{todonotes}

\icmltitlerunning{DiffStitch: Boosting Offline Reinforcement Learning with
Diffusion-based Trajectory Stitching}

\begin{document}

\twocolumn[
\icmltitle{DiffStitch: Boosting Offline Reinforcement Learning with \\ Diffusion-based Trajectory Stitching}




\begin{icmlauthorlist}
\icmlauthor{Guanghe Li}{jlu}
\icmlauthor{Yixiang Shan}{jlu}
\icmlauthor{Zhengbang Zhu}{sjtu}
\icmlauthor{Ting Long}{jlu}
\icmlauthor{Weinan Zhang}{sjtu}
\end{icmlauthorlist}

\icmlaffiliation{jlu}{Jilin University}
\icmlaffiliation{sjtu}{Shanghai Jiao Tong University}


\icmlcorrespondingauthor{Ting Long}{longting@jlu.edu.cn}


\vskip 0.3in
]



\printAffiliationsAndNotice{} % otherwise use the standard text.

\begin{abstract}

In offline reinforcement learning (RL), the performance of the learned policy highly depends on the quality of offline datasets. However, in many cases, the offline dataset contains very limited optimal trajectories, which poses a challenge for offline RL algorithms as agents must acquire the ability to transit to high-reward regions. To address this issue, we introduce Diffusion-based Trajectory Stitching (DiffStitch), a novel diffusion-based data augmentation pipeline that systematically generates stitching transitions between trajectories. DiffStitch effectively connects low-reward trajectories with high-reward trajectories, forming globally optimal trajectories to address the challenges faced by offline RL algorithms. Empirical experiments conducted on D4RL datasets demonstrate the effectiveness of DiffStitch across RL methodologies. Notably, DiffStitch demonstrates substantial enhancements in the performance of one-step methods (IQL), imitation learning methods (TD3+BC), and trajectory optimization methods (DT). 
% Our code is publicly available at 
% \url{https://anonymous.4open.science/r/DiffStitch-6F22}.

\end{abstract}

\section{Introduction}
\label{submission}
% \longting{shall we start the story from offline RL? It's too far to start from RL.}
% Deep reinforcement learning has demonstrated significant success across various domains, such as chess \cite{silver2017mastering}, robot control \cite{hua2021learning}, and large language models \cite{ouyang2022training}. Many of these applications involve interacting with the environment to gather a huge amount of data and learn a policy from the collected data\cite{haarnoja2018soft}. However, in numerous real-world scenarios, conducting a large number of interactions with the environment is both costly and dangerous \cite{fujimoto2021minimalist}. To tackle this issue, offline RL (batch RL) \cite{fu2020d4rl} provides a safe and promising approach, which trains a policy from a fixed dataset, eliminating the necessity for direct interaction with the environment.

Recently, offline reinforcement learning (RL)~\cite{levine2020offline,agarwal2020optimistic,fujimoto2021minimalist,janner2021offline,kidambi2020morel}, which focuses on learning a policy from pre-collected, static datasets without directly interacting with the environment, has gained much attention. 
It has particularly wide application in scenarios where obtaining real-time feedback is costly, time-consuming, or impractical \cite{levine2020offline}.
By learning from the offline dataset, offline RL avoids the necessity of direct interaction with the environment, effectively eliminating the associated costs and risks.

\begin{figure}[t]
    \vskip 0.2in
    \begin{center}
    \centerline{\includegraphics[width=\columnwidth]{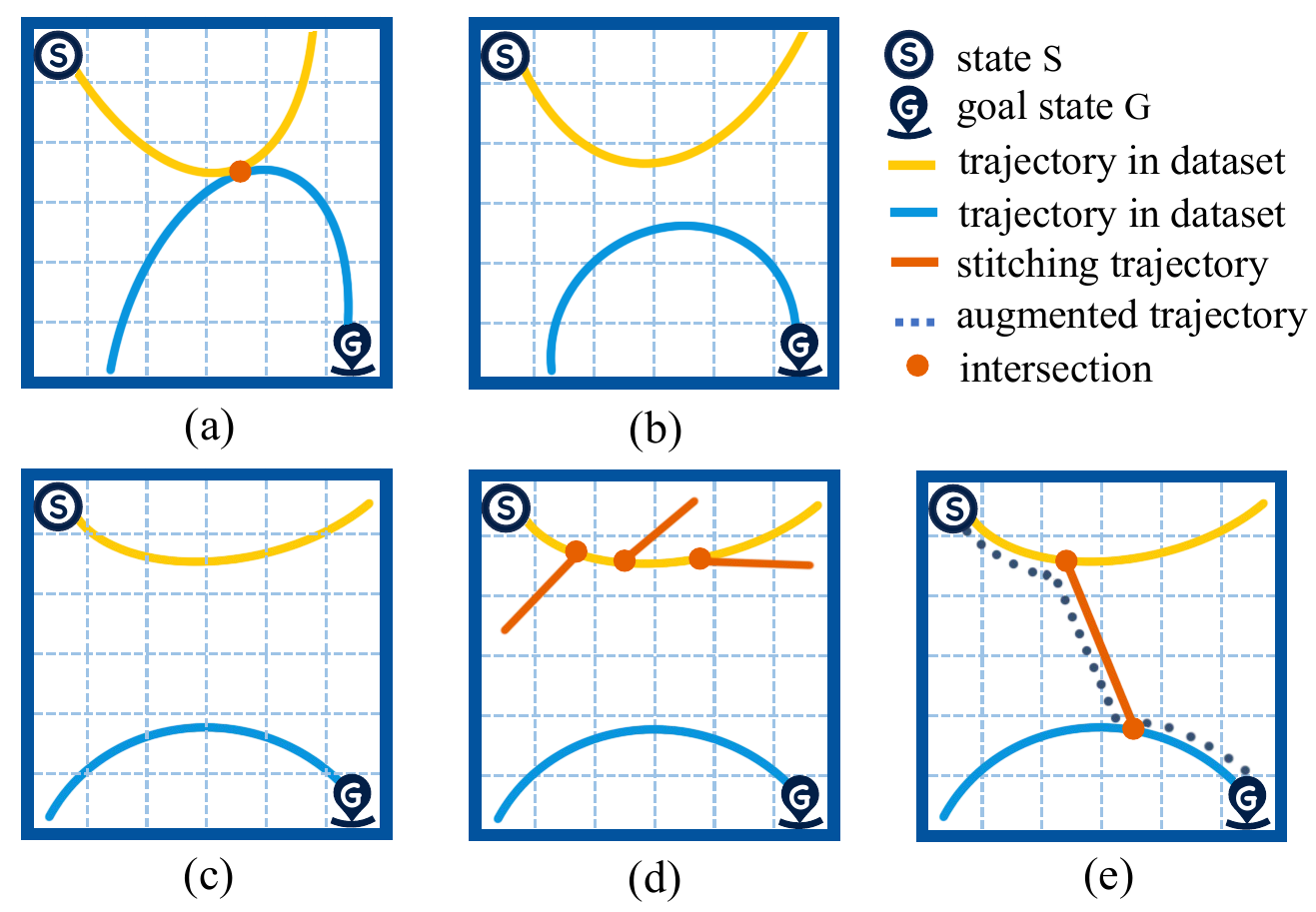}}
    \vspace{-5pt}
      \caption{
      % An illustrative example of why stitching is necessary. 
      An illustration of trajectory stitching.
      Suppose there are two trajectories (blue and yellow) in the offline dataset, and the objective for the agent is to learn a policy that starts from $S$ and reaches $G$. (a) and (b) present the scenarios where the trajectories in the offline dataset intersect or are in close proximity, making it easier to learn a policy that leads to $G$. (c) presents the scenario where trajectories are far apart, posing a challenge for learning a viable policy.
      % to reach $G$.
      (d) illustrates previous solutions that generate trajectories based on original data to enhance policy learning. Although many branches extend from the original trajectory that starts at $S$, none of them formalizes a sample trajectory that starts from $S$ and reaches $G$. 
      (e) illustrates our solution which stitches the trajectory starting from $S$ and ending at $G$, facilitating policy learning by providing a clear path to follow.
      }
      % In all subfigures, the start state $S$ is in the top-left, and the goal state $G$ is in the bottom-right. \textbf{(a)Two trajectories intersect}.  In this case, stitching is unnecessary since there is a trajectory from start to goal. \textbf{(b)Two trajectories are separate and close together}. In this secario, Offline RL algorithms can stitch because neural nets have the ability to generalize. \textbf{(c)Two trajectories are separate and far away. } Offline RL algporithms cannot perform stitching via generalization because they are too far away from each other. \textbf{(d)Two separate trajectories are stitched by additionl transitions}. Additionally adding transitions that connect two trajectories can help offline RL algorithms perform stitching.}
      \vspace{-5pt}
      \label{fig:intro_toy}
    \end{center}
    \vskip -0.2in
    \vspace{-10pt}
\end{figure}

% Despite this, the offline dataset, where an offline RL algorithm learns a policy, has a significant impact on the learning of policy. A low-quality offline dataset usually leads to a worse performance policy.
% Although the offline RL methods have made some achievements in commercial recommendation~\citep{xiao2021general}, health care~\citep{fatemi2022semi}, dialog~\citep{jaques2020human} and autonomous driving~\citep{shi2021offline}, their performance is strongly constrained by the offline dataset.%

While offline RL has achieved notable sucess in commercial recommendation , health care \cite{fatemi2022semi}, dialog \cite{jaques2020human} and autonomous driving \cite{shi2021offline}, its performance is heavily dependent on the quality of the offline dataset. When the offline dataset suffers from inherent deficiencies or shortcomings, the performance of offline RL will remarkably decline. 
% \guanghe{In this paper, we examine a specific shortcoming of offline datasets-the lacking of optimal trajectories } 
% \guanghe{In offline RL, the performance of learned policy is strongly constrained by the quality of offline dataset. When the offline dataset suffers from inherent deficiencies or shortcomings, the performance of offline RL methods can experience a notable decline. In this paper, we focus on tackling the challenge presented by offline datasets that lack globally optimal trajectories.}
% Unfortunately, offline dataset often suffers from inherent deficiencies or shortcomings. 
Consider the task in Figure \ref{fig:intro_toy}, where an agent starts from state $S$ and receives a positive reward only when it reaches the goal state $G$. Suppose there are only two trajectories in the offline dataset, one starting from state $S$, and one ending at state $G$. If the states of two trajectories intersect, as depicted in Figure \ref{fig:intro_toy}(a), 
it is easy for a typical offline RL algorithm such as TD3+BC \cite{fujimoto2021minimalist} to learn a policy that starts from state $S$ and reaches state $G$ through temporal difference learning\cite{sutton2018reinforcement}.
In cases where two trajectories are disjoint but have some very close states, as shown in Figure \ref{fig:intro_toy}(b), learning a policy that starts from $S$ and achieves $G$ is also possible due to the generalization capabilities of value networks. However, in more challenging cases where two trajectories are disjoint and distant from each other as illustrated in Figure \ref{fig:intro_toy}(c), learning a policy that starts from $S$ and achieves $G$ becomes much more difficult.

To deal with the aforementioned challenges of data deficiency, a straightforward solution is augmenting the dataset: generating synthetic data to facilitate learning a better policy. 
Therefore, previous works \cite{lu2023synthetic, zhang2023uncertainty,wang2022bootstrapped} propose to learn the state transition of offline datasets and generate relatively short sub-trajectories from a randomly selected state, as Figure \ref{fig:intro_toy}(d) illustrated. However, without specifying a desired target state, the generated sub-trajectories may not effectively enhance the policy to achieve a higher return. As shown in Figure \ref{fig:intro_toy}(d), although many branches extend from the original trajectory that starts at $S$, none of them formalize a sample trajectory that starts from $S$ and reaches $G$, learning the policy of reaching $G$ is still difficult.
% are difficult to connect to the trajectory ends at $G$. In this case, the augmented data composed by the generated sub-trajectory has no benefit for policy learning. 
Intuitively, if we augment the offline dataset by stitching together low-reward trajectories with high-reward trajectories, the policy may learn the ability to transit from low-reward states to high-reward states and ultimately achieve a higher overall return. For instance, if we generate a sub-trajectory to connect (stitch) the trajectory that starts from $S$ and the trajectory that ends at $G$ as Figure \ref{fig:intro_toy}(e), the new trajectory (the dashed line in Figure \ref{fig:intro_toy}(e)) will facilitate the policy learning. 
% \longting{learning the path from $s$ to $g$ is much difficult. Fortunately, stitching the sub-trajectories in the offline dataset has significant potential to address the limitations, as Figure \ref{fig:intro_toy}(d) illustrates. The stitching trajectory demonstrates a path from $s$ to $g$, which facilitates the learning of offline RL algorithms. }
% stitching becomes possible only if additional stitching transitions are generated to connect them.

% In summary, the absence of globally optimal trajectories makes offline RL harder compared to online RL. Enhancing dataset quality is equally crucial alongside exploring novel offline RL algorithms. 

% \longting{Discuss previous stitching works, and their limitations.}

Considering that, in this paper, we propose Diffusion-based Trajectory Stitching(DiffStitch), 
% which is composed of four modules and 
a novel data augmentation method designed to stitch (connect) the low-reward trajectories with high-reward trajectories in offline dataset to enhance the learning of policies. Specifically, it randomly selects a low-reward trajectory and a high-reward trajectory, and generates a sub-trajectory to stitch them together.
% . Then, it estimates the number of required steps to connect the two trajectories with the step estimation module. Next, it generates a state sequence to stitch the states of the low-reward trajectory and the high-reward trajectory with the state stitching module. Subsequently, it wrap-up the states with rewards and actions to obtain a sub-trajectory in the wrap-up module. Finally, the sub-trajectory is connected with the low-reward and high-reward sub-trajectories to generate the new trajectory, qualified in the qualification module. If the qualification is satisfied, the new trajectory is put into the augmented data.
% In this paper, we present DiffStitch, a novel paradigm that generates transitions via a generative model and a rule-based discriminator that eliminates low-quality generations. This paradigm is capable of smoothly stitching together any two separate trajectories, which enhances the original offline dataset by covering more states and connecting different local optimal regions. Different from prior methods that only use forward or backward generation, we propose a bi-directional generating mechanism. The mechanism leverages a diffusion model to autonomously generate transitional states between two trajectories. 
Through our paradigm, one can easily transfer the trajectories trapped in low rewards to the one with high rewards, enhancing the learning of policy and significantly improving the performance of offline RL algorithms.% one can easily generate extra transitions that transform a suboptimal dataset into an optimal dataset, containing global optimal trajectories, significantly improving the final performance of an offline RL agent.
To the best of our knowledge, DiffStitch is the first offline RL augmentation method that generates sub-trajectories to stitch any two trajectories in the dataset. We evaluate DiffStitch on various offline datasets, and results demonstrate that augmented trajectories generated by DiffStitch are effective in enhancing the performance of different types of offline RL algorithms.

% Our approach enhances various offline RL methods, including Behavior Cloning, Bellman function-based algorithms, and sequential modeling approaches.

Our contributions are summarized as follows:
% \longting{Four contributions are too many. Try to change four to three.}
% \zhengbang{I think the last one can be removed, or integrated with the first one.}
\begin{itemize}
\item We propose DiffStitch, a novel paradigm for augmenting the offline RL dataset by trajectory stitching. To the best of our knowledge, DiffStitch is the first offline RL augmentation method that generates sub-trajectories to stitch any two trajectories in the dataset. 
\item DiffStitch benefits the performance of various offline RL algorithms, including one-step methods(IQL), imitation learning methods(TD3+BC), and sequential optimization methods(DT).  
\item The extensive experiments on the widely-used D4RL datasets demonstrate the superiority of our method.
% BC, Bellman function-based methods, and sequential decision modeling methods. 
% to enhance the performance of offline RL algorithms by stitching the trajectories in the offline RL dataset.
% \item We propose a novel framework designed to enhance an offline RL dataset by generating stitched transitions between any two trajectories. This augmentation provides the offline RL agent with stitched demonstrations, thereby alleviating the challenges of offline learning. This allows future researchers to explore and discover the selection of concatenation paths on the basis of the framework we have provided.   
% \item Our method can enhance various downstream offline RL algorithms, including BC, Bellman function-based methods, and sequential decision modeling methods. 
% \item Through extensive experiments on the widely-used offline RL dataset D4RL, we demonstrate the superiority of our method.
\end{itemize}

\section{Related Work}

% \begin{figure}[ht]
%     \vskip 0.2in
%     \begin{center}
%     \centerline{\includegraphics[width=\columnwidth]{figs/intro_toy_2.pdf}}
%       \caption{An illustrative example of why stitching is necessary. In all subfigures, start state is in top-left corner, and goal state is in down-right corner. \textbf{(a)Two trajectories intersect}.  In this case, stitching is uneccessary since there exits a trajectory from start to goal. \textbf{(b)Two trajectories are separate and close together}. In this secario, Offline RL algorithms can stitch because neural nets have the ability to generalize. \textbf{(c)Two trajectories are separate and far away. } Offline RL algporithms cannot perform stitching via generalization because they are too far away from each other. \textbf{(d)Two separate trajectories are stitched by additionl transitions}. Additionally adding transitions that connect two trajectories can help offline RL algorithms perform stitching.}
%       \label{fig:intro_toy}
%     \end{center}
%     \vskip -0.2in
% \end{figure}

\textbf{Offline RL.} 
According to \cite{prudencio2023survey}, offline RL algorithms could be grouped into four groups: model-based methods, one-step methods, imitation learning, and trajectory optimization methods. 
Model-based methods~\cite{kidambi2020morel,matsushima2020deployment,yu2020mopo} learn policy by using the offline dataset to model the dynamic of the environment.
% For instance, \longting{MOPO \cite{yu2020mopo} learns the dynamic of the environment first, and uses the disagreements in dynamics models to constrain the learning.}
One-step methods \cite{kostrikov2021offline,brandfonbrener2021offline} learn policy by performing in-sample Bellman updates to learn an accurate estimate of the Q function, which is followed by a single policy improvement step to find the best possible policy. 
% For instance, IQL \cite{kostrikov2021offline} first learns the value function and then derives the policy based on the learned value function.
Imitation learning methods \cite{chen2020bail,wang2020critic,siegel2020keep} learn the policy by mimicking the optimal behaviors and filtering out suboptimal behaviors.
% For instance, \longting{TD3+BC takes advantage of deep deterministic policy gradient (TD3) to enhance behavior cloning (BC).}
Trajectory optimization \cite{chen2021decision,janner2021offline} methods focus on training a joint state-action model over entire trajectories. 
% For instance, DT \cite{chen2021decision} is designed to leverage the Transformer \cite{vaswani2017attention} to model the distribution of trajectories, and minimize the mean squared error (MSE) between the predicted and ground-truth actions from the trajectory.
Although these offline RL methods have made significant achievements in recent years, their performance is highly dependent on the offline dataset.

\begin{figure*}[t]
  \centering
  \includegraphics[width=1\textwidth]{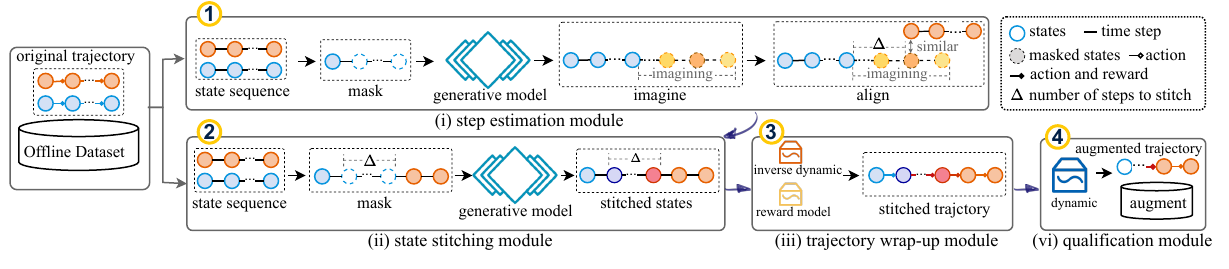}
    \vspace{-20pt}
  \caption{The overall pipeline framework of DiffStitch.}
  \vspace{-15pt}
  \label{fig:pipeline}
\end{figure*}

\textbf{Data Augmentation in Offline RL.} 
Some of the previous works propose to enhance the performance of offline RL algorithms by data augmentation.
% , as the performance of offline RL algorithms is highly dependent on the offline dataset. 
We group these methods into single-way rollout \cite{wang2021offline,lyu2022double,zhang2023uncertainty} and dynamic tuple simulation \cite{lu2023synthetic}. The single-way rollout methods design (inverse) dynamic models and a rollout policy to synthesize the augmented trajectories.
For instance, TATU \cite{zhang2023uncertainty} employs a forward dynamic model to roll out trajectories and adopt a truncates mechanism to guarantee the accumulated certainty. The dynamic tuple simulation methods apply generative models to simulate the environment dynamics and generate new transition tuples. For instance, SER \cite{lu2023synthetic} applies diffusion models to capture the joint distribution of transition tuples in the offline dataset and augment the dataset by sampling new tuples.
Although these methods did enhance the performance of offline RL, 
% they are limited in some challenging scenarios, like generating long trajectories and learning from limited high-reward trajectories.
they are incapable of generating long trajectories and are limited in challenging scenarios where very few high-reward trajectories are available in the original offline dataset.

\section{Preliminaries}

Consider a standard Markov Decision Process (MDP) $\langle \mathcal{S}, \mathcal{A}, \rho_{0}, p, r, \gamma \rangle$, where $\mathcal{S}$ denotes the state space, $\mathcal{A}$ denotes the action space, $\rho_{0}$ denotes the initial state distribution, $p$ denotes the transition function, $\gamma$ represents the discount factor, and $r$ is the instant reward. At each time step $t$, an agent takes action $\bm{a}_t (\bm{a}_t \in \mathcal{A})$ to respond to the state of environment $\textbf{s}_t (\bm{s}_t \in \mathcal{S})$ according to a policy $\pi(\bm{a}_t|\bm{s}_t)$, and gains an instant reward $r_t$. Then the time step transfers to $t + 1$, and the state of the environment transits to $\bm{s}_{t+1}$ with transition probability $p(\bm{s}_{t+1}|\bm{s}_t, \bm{a}_t)$. 
% Suppose the total time steps of the agent interacting with the environment is $T$, 
The interaction between the agent and environment could be written as a trajectory $\tau=\{(\bm{s}_{1}, \bm{a}_{1}, r_{1}), (\bm{s}_{2}, \bm{a}_{2}, r_{2}), \dots\}$.
% in which $\bm{s}_t, \bm{a}_t, r_t$ denote the state, action and reward at step $t$ correspondingly.
The goal of the agent is to learn a policy $\pi^*$ that can maximize expected discounted return, \textit{i.e.}, $\sum_{t=0}^{\infty}\gamma^{t}r_t$.

% Since the agent may not be allowed to directly interact with the environment in some scenarios, 
In the setting of offline RL, given an offline dataset $\mathcal{D}$ which is composed of pre-collected trajectories, the agent is supposed to learn an optimal policy from $\mathcal{D}$ (no further interactions are allowed).
Hence, for an arbitrary offline RL algorithm $\Gamma$, it is first trained with the offline dataset $\mathcal{D}$ to obtain the optimal policy $\pi$, and we have $\pi = \Gamma(\mathcal{D})$. Subsequently $\Gamma$ is tested in the environment to evaluate its performance. Specifically, the performance is computed by:
% Given an offline RL algorithm $\Gamma$ that takes as input an offline dataset and return a learned policy $\pi=\Gamma(D)$, the performance of the policy $\pi$ can be written as:
\begin{equation}\label{eq:off_tar}
    \mathbb{E}_{\bm{\tau}\sim\Gamma(\mathcal{D})} \left [\sum_{t=0}^{\infty}\gamma^{t}r_t\right]~.
\end{equation}
Previous works mainly focused on maximizing Eq.~(\ref{eq:off_tar}) by designing a better learning algorithm $\Gamma$. 
Obviously, the performance of policy $\pi$ also largely depends on the quality of offline dataset $\mathcal{D}$.
% according to Eq.~(\ref{eq:off_tar}),
% and enhancing the quality of $\mathcal{D}$ benefits the training of $\pi$. 
Therefore, in this paper, we aim to generate an augmented dataset $\mathcal{D}^*$, such that for any algorithm $\Gamma$ in a set of reasonable learning algorithms, the policy learned from $\mathcal{D}^*$ would achieve better performances compared to learning from $\mathcal{D}$:
\begin{equation}
     \mathbb{E}_{\bm{\tau}\sim\Gamma(\mathcal{D})} \left [\sum_{t=0}^{\infty}\gamma^{t}r_t \right ] < \mathbb{E}_{\bm{\tau}\sim\Gamma(\mathcal{D}^*)} \left [\sum_{t=0}^{\infty}\gamma^{t}r_t \right ]~.
     % \mathbb{E}_{\bm{\tau}\sim\Gamma(\mathcal{D})} \left [\sum_{t=0}^{\infty}\gamma^{t}r(s_{t}, a_{t})\right ] < \mathbb{E}_{\bm{\tau}\sim\Gamma(\mathcal{D}^*)} \left [\sum_{t=0}^{\infty}\gamma^{t}r(s_{t}, a_{t}) \right ]~.
\end{equation}

% Since the offline dataset, $\mathcal{D}$ might have the problem of lack of diversity and mixed quality \cite{}, we aim to address the offline RL issue by data augmentation. Specifically, we such that policies learned on $\mathcal{D}^*$ in this paper, such that policies learned on would achieve better performances compared to learning on $\mathcal{D}$.

%  Thus, if we are able to enhance $D$ into $D^*$, \longting{we need to give the superiority of $D^*$, such that it could enhance the performance}. then the final performance of the learned policy would likely to be improved.

% Thus, we aim to generate $D_{aug}$ from $D$ in this paper, and merge them together to form an enhanced dataset $D^*$, such that policies learned on $D^*$ would achieve better performances compared to learning on $D$.

% \longting{marking}
% \begin{equation}
%     D_{aug} = \textbf{DiffStitch}(D)
% \end{equation}

 % \longting{in training. (Briefly describe the limitation of this setting). To achieve the goal defined in Eq. xx (previously defined equation), we transfer Eq.xxx as: (give the goal of the equation of this paper and give a detailed explanation of why the new goal could address the problem). remove the following sentences}. To achieve an optimal policy, an offline RL agent must identify optimal regions in the dataset $D$ and attempt to stitch them together, which is a challenging task. In the following section, we focus on mitigating this issue by generating stitching transitions that help combine different optimal regions to enhance offline learning.

\section{Method} 
% To enhance $D$, 
To obtain the augmented dataset $\mathcal{D}^*$, we propose DiffStitch, solving the problem 
% \longting{by stitching the trajectories with low reward to the trajectories with high reward.} 
from the perspective of trajectory stitching.
In the following, we will discuss the detailed pipeline of DiffStitch and subsequently the implementation and training of the models involved. 

\subsection{DiffStitch} \label{sec:diffstitch}
% In Stitching Module, w
We aim to augment the offline dataset $\mathcal{D}$ by stitching, which generates the transitions that ``link'' low-reward trajectories to high-reward trajectories, as illustrated in Figure \ref{fig:intro_toy}(e).
% Given low-reward trajectory $\tau=\{(\bm{s}_{1}, \bm{a}_{1}, r_{1}),\\ (\bm{s}_{2}, \bm{a}_{2}, r_{2}), \dots, (\bm{s}_{T}, \bm{a}_{T}, r_{T})\}$ and high-reward trajectory $\tau'=\{(\bm{s}'_{1}, \bm{a}'_{1}, r'_{1}), (\bm{s}'_{2}, \bm{a}'_{2}, r'_{2}), \dots, (\bm{s}_{T'}', \bm{a}_{T'}', r_{T'}')\}$, 
Suppose there is a low-reward trajectory $\tau=\{(\bm{s}_{1}, \bm{a}_{1}, r_{1}), \dots, (\bm{s}_{T}, \bm{a}_{T}, r_{T})\}$ and a high-reward trajectory $\tau'=\{(\bm{s}'_{1}, \bm{a}'_{1}, r'_{1}), \dots, (\bm{s}_{T'}', \bm{a}_{T'}', r_{T'}')\}$, 
where $T$ and $T'$ denote the length of $\tau$ and $\tau'$. A straightforward way to stitch them together is directly feeding a masked trajectory composed of the elements of $\tau$ and $\tau'$, \textit{i.e.}, $\{(\bm{s}_{1}, \bm{a}_{1}, r_{1}), \dots, (\bm{s}_{T}, \bm{a}_{T}, r_{T}),  \texttt{[MASK]}, ..., \texttt{[MASK]},\\ (\bm{s}'_{1}, \bm{a}'_{1}, r'_{1}), ..., (\bm{s}'_{T'}, \bm{a}'_{T'}, r'_{T'})\}$, to a generative model and predict the \texttt{[MASK]}.
% Here we consider using diffusion models as the generative model since they are successful in  similar inpainting tasks among various domains~\cite{lugmayr2022repaint,tevet2022human}.
However, this naive approach has a significant drawback in that it can not determine the number of \texttt{[MASK]} ( \textit{i.e.}, the time steps lying between $\tau$ and $\tau'$). 
% Using a fixed time step between two trajectories could result in values that are either too large or too small. 
A fixed time step inserted between two trajectories could be either too large or too small, which leads to a violation of environmental dynamics and resides in out-of-distribution (OOD) regions. Besides, rewards and actions usually do not change smoothly as states across time, so generating them with a single model might bring some potential issues \cite{ajay2022conditional}.

To tackle this problem, our proposed DiffStitch first estimates the steps for stitching and then generates the states, actions, and rewards separately. Specifically, as illustrated in Figure \ref{fig:pipeline}, it is composed of four modules: step estimation module, state stitching module, trajectory wrap-up module, and qualification module. The step estimation module estimates the number of time steps required for stitching two trajectories. Then, the estimated number of time steps is input into the state stitching module, which generates states consistent with the number of estimated steps. Next, the trajectory wrap-up module predicts the rewards and actions based on the states generated by the state stitching module to obtain a new trajectory. Finally, the qualification module evaluates the quality of the new trajectory and augments the dataset with the new trajectory if it is qualified.

% \longting{This problem is analogous to calculating the number of stops between a specific bus station and a destination. When we want to identify the number of stops between them,  we typically start from that bus station and find the nearest bus station along the route to the destination. The count of stops between two bus stations is the number of stops between a specific bus station and a destination. Inspired by this problem, in order to determine the time steps between two states, }

% \guanghe{When individuals are tasked with navigating  from A to B, their performance is likely to improve when they are aware of how to navigate from A to B', where B' is close to B. Drawing inspiration from this concept, in order to determine the number of steps for connecting $\bm{s}_T$ to $\bm{s}'_1$, we initially 'imagine' the subsequent states of state $\bm{s}_T$—states for which the navigation is well-understood. We then identify states with high similarity to $\bm{s}'_1$, treating the intervals between $\bm{s}_T$ and the state close to $\bm{s}'_1$ as the number of required step.}

\subsubsection{Step Estimation Module}
As discussed in previous work \cite{ajay2022conditional} that the state transitions are smooth within a trajectory, we use states to estimate the number of steps for stitching. Therefore, the task of step estimation is transformed that given the states of two trajectory $\tau_{s}=(s_{1},s_{2}...s_{T})$ and $\tau_{s}'=(s'_{1},s'_{2},...,s'_{T'})$, estimate the number of the step required for state $\bm{s}_T$ to transit naturally to state $\bm{s}'_1$.
% \textbf{Imagination}. In this stage, we aim to identify the number of steps lying between $\bm{s}_T$ and $\bm{s}'_1$, where 
Here, $\bm{s}_T$ could be viewed as the start state and $\bm{s}'_1$ could view as the target state for stitching. 
Suppose $\bm{s}_T$ can reach state $x$ in exactly $i$ steps. The higher the similarity between $\bm{s}'_1$ and $x$, the more probable it is for $\bm{s}_T$ to reach $\bm{s}'_1$ in exactly $i$ steps as well. Inspired by that, to identify the number of steps required for stitching $\bm{s}_{T}$ to $\bm{s}'_1$, we first “imagine” the subsequent states of state $\bm{s}_{T}$, find the state closest to $\bm{s}'_1$, and treating the interval between $\bm{s}_{T}$ and the state closest to $\bm{s}'_{1}$ as the number of required step.

% Motivated by this intuition, to identify the number of steps needed to transit from $\bm{s}_T$ to $\bm{s}'_1$, we first ``imagine'' the subsequent states of state $\bm{s}_T$, denoted as $(\bm{s}_{1}^m, ..., \bm{s}_{H-1}^m)$. For each state $s^{m}_{i}$, it is apparent that $s_{T}$ can reach $s^{m}_{i}$ in $i$ steps. We then identify the state $s^m_{i^*}$ that is most similar to $\bm{s}'_1$. The value of $i^*$ can be interpreted as the most probable number of steps required to transition between the two trajectories.  

% compare the similarities of the subsequent state and $\bm{s}'_1$, and set the states with highest similarity as the closet one.

Specifically, we first initialize a masked sequence starting with state $\bm{s}_T$, \textit{i.e.}, $(\bm{s}_T, \texttt{[MASK]}, ..., \texttt{[MASK]})$, and feed it to a conditional generative model to generate the subsequent states of $\bm{s}_T$:
\begin{equation} \label{eq:image}
    \begin{aligned}
        \tau_{s}^m &=(\bm{s}_{T}, \bm{s}_{1}^m, ..., \bm{s}_{H-1}^m), \\
        &=\mathcal{G}_{\theta} \left((\bm{s}_T, \texttt{[MASK]}, ..., \texttt{[MASK]})\right),
        % \widetilde{\tau_{s}} &=(\bm{s}_{T}, \widetilde{\bm{s}_{1}}, ..., \widetilde{\bm{s}_{H-1}}), \\
        % &=\mathcal{G}_{\theta}(\bm{s}_T, \texttt{[MASK]}, ..., \texttt{[MASK]}),
    \end{aligned}
\end{equation}
% where $H$ is the input horizon of generative model 
where $\mathcal{G}_{\theta}$ denotes the generative model,
% we implement it with diffusion model \cite{} \longting{todo}, and 
the implementation and training of $\mathcal{G}_{\theta}$ will be discussed in Section \ref{sec:train_learn}. $H$ is the generating horizon of the generative model $\mathcal{G}_{\theta}$. 
% $(\bm{s}_T, \texttt{[MASK]}, ..., \texttt{[MASK]})$ denotes a matrix, in which all rows except the row represents $\bm{s}_T$ are set to Gaussian noise. With many denoise steps of diffusion, we obtain the ``imaged'' trajectory $\tau_{s}^m$. Therefore, the subsequent states of $\bm{s}_T$ are $(\bm{s}_{1}^m, ..., \bm{s}_{H-1}^m)$. 
It is worth noting that, $\bm{s}_T$ is considered as the only condition in Eq.~(\ref{eq:image}). 

% $\bm{s}'_1$ is not used as conditions here, as it is currently unknown which steps $\bm{s}'_1$ should be located. \longting{remove this sentence}

% as which steps $\bm{s}'_1$ should be shown is currently unknown.

Next, we compare the similarity of $\textbf{s}'_{1}$ and the subsequent states of $\bm{s}_{T}$, \textit{i.e.}, $(\bm{s}_{1}^m, ..., \bm{s}_{H-1}^m)$, to identify the number of steps required to transit between two trajectories, which could be calculated by:
\begin{equation}
    \Delta = \argmax_{i} \text{sim}(\bm{s}_{i}^m, \bm{s}'_{1}),
\end{equation}

% whether the two trajectories are stitchable. This can be judged by the following condition: 

% \begin{equation}
%     \max_{i} \{ \text{sim}(s_{i}^{m}, s'_{1})\} \ge \alpha 
% \label{condition-alpha}
% \end{equation}

% Here, $\alpha$ is the similarity threshold. 
% % \sout{If the minimum distance between $\bm{s}_{i}^m$ and $\textbf{s}'_{1}$ is less than a similarity threshold $\alpha$, it indicates that the number of steps between $\textbf{s}_{T}$ and $\textbf{s}'_{1}$ could be identified by:}
% If the condition in equation \ref{condition-alpha} holds, it indicates that the two trajectories is "stitchable", 

% and the number of steps between $\textbf{s}_{T}$ and $\textbf{s}'_{1}$ 

where $\Delta$ denotes the number of time steps to connect $\bm{s}_T$ and $\bm{s}'_1$. Obviously, $\Delta \leq H - 2$ . 
Here, we implement $\text{sim}(\cdot)$ with cosine similarity. 
\subsubsection{State Stitching Module}
% \subsubsection{Stitching}
With $\Delta$, we are able to know the exact number of states to be generated between the last state of trajectory $\tau=\{(\bm{s}_{1}, \bm{a}_{1}, r_{1}), (\bm{s}_{2}, \bm{a}_{2}, r_{2}), \dots, (\bm{s}_{T}, \bm{a}_{T}, r_{T})\}$ and the first state of trajectory $\tau'=\{(\bm{s}'_{1}, \bm{a}'_{1}, r'_{1}), (\bm{s}'_{2}, \bm{a}'_{2}, r'_{2}), \dots, \\
(\bm{s}_{T'}', \bm{a}_{T'}', r_{T'}')\}$.
% $\tau_{s}=(s_{1},s_{2}...s_{T})$ and $\tau_{s}'=(s'_{1},s'_{2},...,s'_{T'})$. 
Applying this information,  we first construct a masked sequence composed of the last states of $\tau$ and the initial states in $\tau'_s$:
% create a condition mask as follows: 
\begin{equation} \label{eq:s_mat}
\begin{aligned}
    % &\textbf{M}_s =
    \tau_{s,m} = (\bm{s}_{T}, \underbrace{\texttt{[MASK]}, \ldots, \texttt{[MASK]}}_{\Delta \text{ masks}} , \bm{s}'_{1}, ..., \bm{s}'_{H-1-\Delta}),
    % &\text{MASK$(\tau_{s}, \tau'_{s})$} =(s_{T}, \underbrace{\texttt{[MASK]}, \ldots, \texttt{[MASK]}}_{\Delta \text{ Masks}} , s'_{1}, ..., s'_{H-1-\Delta})   
\end{aligned}
\end{equation}

% Here, $H$ is the input horizon of generative model $\mathcal{G}_{\theta}$.
Here, the required elements of mask sequence $\tau_{s,m}$ are state $\bm{s}_T$ and and $\bm{s}'_1$. $(\bm{s}'_{2}, \bm{s}'_{3}, ..., \bm{s}'_{H-1-\Delta})$ is used to pad the sequence if $\Delta < H - 2$.
% Since $\Delta+2 \leqslant H$ always holds, we can just patch the rest of the mask with $(s'_{2}, s'_{3}, ..., s'_{H-1-\Delta})$ if necessary.
Then, we apply $\mathcal{G}_{\theta}$ to reconstruct the masked elements, and states that stitch $\bm{s}_T$ to $\bm{s}'_1$ can be generated as follows: 
\begin{equation} \label{eq:tauss}
\begin{aligned}
    \tau_s^s &= \mathcal{G}_{\theta}(\tau_{s,m}) \\ 
    &= (\bm{s}_{T}, \underbrace{\widetilde{s_{1}}, \ldots, \widetilde{\bm{s}_{\Delta}}}_{\Delta \text{ stitching states}} , \bm{s}'_{1}, ..., \bm{s}'_{H-1-\Delta}), 
    % \text{Stitch}(\tau_{s},\tau'_{s})&=\mathcal{G}_{\theta}(\text{MASK}(\tau_{s}, \tau'_{s})) \\ 
    % &= (s_{T}, \underbrace{\widetilde{s_{1}}, \ldots, \widetilde{s_{\Delta}}}_{\Delta \text{ transition states}} , s'_{1}, ..., s'_{H-1-\Delta})
\end{aligned}
\end{equation}
Here, we denote the states $\widetilde{\tau_s} = (\widetilde{\bm{s}_{1}}, \ldots, \widetilde{\bm{s}_{\Delta}})$ are the \textbf{stitching states}.

\subsubsection{Trajectory Wrap-up Module} \label{sec:wrap-up}
% With the stitching states $\widetilde{\tau_s} = (\widetilde{\bm{s}_{1}}, \ldots, \widetilde{\bm{s}_{\Delta}})$ generated by Eq.~\ref{eq:tauss}, w
We aim to wrap-up the states with actions and rewards to obtain the sub-trajectory for stitching.
% Following \cite{ajay2022conditional}, 
Specifically, for each state pair $(\hat{\bm{s}_t}, \hat{\bm{s}}_{t+1}) \in \{(\bm{s}_T,\widetilde{\bm{s}_{1}}), (\widetilde{\bm{s}_{1}}, \widetilde{\bm{s}_{2}}),\ldots, (\widetilde{\bm{s}_{\Delta}}, \bm{s}'_1)\}$ , we apply a inverse dynamic model $f_\psi$ to predict the actions between two adjacent states, that is:
% and each state $\hat{\bm{s}}_{t+1} \in \{\widetilde{\bm{s}_{1}}, \ldots, \widetilde{\bm{s}_{\Delta}}, \bm{s}'_1\}$, 
\begin{equation} \label{eq:action}
    % \widetilde{\bm{a}_t} = f_{\psi}(\widetilde{\bm{s}_t}, \widetilde{\bm{s}_{t+1}}),
    \widetilde{\bm{a}_t} = f_{\psi}(\hat{\bm{s}}_t, \hat{\bm{s}}_{t+1})
\end{equation}
For each state-action pair $(\hat{\bm{s}_t}, \widetilde{\bm{a}}_{t}) \in \{(\bm{s}_T,\widetilde{\bm{a}_{T}}), (\widetilde{\bm{s}_{1}}, \widetilde{\bm{a}_{1}}), \\ \ldots, (\widetilde{\bm{s}_{\Delta}}, \widetilde{\bm{a}_{\Delta}})\}$, we apply a reward model $f_{\phi}$ to predict the reward between two adjacent states:
\begin{equation} \label{eq:reward}
    \widetilde{r_t} =  f_{\phi}(\hat{\bm{s}}_{t}, \widetilde{\bm{a}_t}), 
\end{equation} 
The implementation and training of $f_\psi$ and $f_{\phi}$ will be discussed in Section \ref{sec:train_learn}.
% It is important to note that the diffusion model $\mathcal{G}_{\theta}$ exclusively generates stitching states. Similar to \cite{ajay2022conditional}, we generate actions through the inverse dynamics model $f_{\psi}(s_{t}, a_{t})$ and generate rewards using the dynamics model $p_{\phi}(r_{t}|s_{t},s_{t+1})$. 

Therefore, the sub-trajectory for stitching can be represented as 
$\tau_{r}=\{(\bm{s}_{T}, \widetilde{\bm{a}_{T}}, \widetilde{r_{T}}), (\widetilde{\bm{s}_{1}}, \widetilde{\bm{a}_{1}}, \widetilde{r_{1}}), ..., (\widetilde{\bm{s}_{\Delta}}, \widetilde{\bm{a}_{\Delta}}, \widetilde{r_{\Delta}}), (\bm{s}'_1,\\ \bm{a}'_1, r'_{1})\}$.  We denote $\tau_{r}$ as the \textbf{stitching trajectory}.

% Therefore, the stitching trajectory can be 
% represented as 
% $\tau_{gen}=\{(\bm{s}_{1}, \bm{a}_{1}, r_{1}), ... , (\bm{s}_{T}, \widetilde{\bm{a}_{T}}, \widetilde{r_{T}}), (\widetilde{\bm{s}_{1}}, \widetilde{\bm{a}_{1}}, \widetilde{r_{1}}), ..., (\\ \widetilde{\bm{s}_{\Delta}}, \widetilde{\bm{a}_{\Delta}}, \widetilde{r_{\Delta}}), (\bm{s}'_{1}, \bm{a}'_{1}, r'_{1}),...,(\bm{s}'_{H-1-\Delta}, \bm{a}'_{H-1-\Delta}, r'_{H-1-\Delta})\}$.   

% \subsection{Validation Module} 
\subsubsection{Qualification Module}
Though the stitching trajectories could be generated with the techniques mentioned previously, their quality often varies. Therefore, we aim to 
% After generating transitions $\tau_{gen}$ that connect two trajectories $\tau_{s}$ and $\tau'_{s}$, we 
assess the quality of generated trajectories, filtering out low-quality data while retaining high-quality data to ensure that the stitching trajectories align with environmental dynamics. 

For the stitching trajectory $\tau_{r}=\{(\bm{s}_{T}, \widetilde{\bm{a}_{T}}, \widetilde{r_{T}}), (\widetilde{\bm{s}_{1}}, \widetilde{\bm{a}_{1}}, \\ \widetilde{r_{1}}), ..., (\widetilde{\bm{s}_{\Delta}}, \widetilde{\bm{a}_{\Delta}}, \widetilde{r_{\Delta}}),(\bm{s}'_1, \bm{a}'_1, r'_{1})\}$, 
% $\tau_{gen}=\{(\bm{s}_{1}, \bm{a}_{1}, r_{1}), ... , (\bm{s}_{T}, \widetilde{\bm{a}_{T}}, \widetilde{r_{T}}), \\ (\widetilde{\bm{s}_{1}}, \widetilde{\bm{a}_{1}}, \widetilde{r_{1}}), ..., (\widetilde{\bm{s}_{\Delta}},\widetilde{\bm{a}_{\Delta}}, \widetilde{r_{\Delta}}),  (\bm{s}'_{1}, \bm{a}'_{1}, r'_{1}),..., (\bm{s}'_{H-1-\Delta}, \\ \bm{a}'_{H-1-\Delta}, r'_{H-1-\Delta})\}$,
we use the dynamics model \cite{liang2023adaptdiffuser} $f_{\omega}$ to assess the consistency of the generated data with the environmental dynamics. Specifically, 
for each tuple $(\hat{\bm{s}}_t, \hat{\bm{a}}_t, \hat{\bm{s}}_{t+1}) \in \{
(\bm{s}_{T}, \widetilde{\bm{a}_{T}}, \widetilde{\bm{s}_{1}}), (\widetilde{\bm{s}_{1}}, \widetilde{\bm{a}_{1}}, \widetilde{\bm{s}_{2}}), ..., (\widetilde{\bm{s}_{\Delta}}, \widetilde{\bm{a}_{\Delta}}, \bm{s}'_{1})\}$, 
% and given a state $\widetilde{\bm{s}_t}$ and action $\widetilde{\bm{a}_t}$ in $\tau_{gen}$,
we predict the next state via:
\begin{equation}
    \hat{\bm{s}}_{t+1}^q = f_{\omega}(\hat{\bm{s}}_t, \hat{\bm{a}}_t).
\end{equation}
% Following a similar approach in \cite{liang2023adaptdiffuser}, for each $(s_{t}, a_{t})$ in $\tau_{gen}$, we utilize the dynamics model $p_{\phi}(\cdot|s_{t},a_{t})$ to predict $\widetilde{s_{t+1}}\sim p_{\phi}(\cdot|s_{t}, a_{t})$
The implementation and training of $f_w$ will be discussed in Section \ref{sec:train_learn}. We discard stitching trajectory $\tau_{r}$ if there is a predicted state $\hat{\bm{s}}_{t+1}$ that deviates significantly from the corresponding state in $\tau_r$, i.e. 
\begin{equation}
    ||\hat{\bm{s}}_{t+1}^q - \hat{\bm{s}}_{t+1}||^{2} \ge \delta,
\end{equation}
where $\delta$ denotes the qualification threshold. This ensures that the stitching trajectory is aligned with the environmental dynamics.

% avoiding OOD data that degrades the learning of downsdream offline rl algorithms. 

By employing the pipeline above, we can obtain a new augmented trajectory $\tau_{gen}=\{(\bm{s}_{1}, \bm{a}_{1}, r_{1}), ... , (\bm{s}_{T}, \widetilde{\bm{a}_{T}}, \widetilde{r_{T}}), (\widetilde{\bm{s}_{1}}, \widetilde{\bm{a}_{1}}, \widetilde{r_{1}}), ..., (\widetilde{\bm{s}_{\Delta}},  \widetilde{\bm{a}_{\Delta}}, \\ \widetilde{r_{\Delta}}), (\bm{s}'_{1}, \bm{a}'_{1}, r'_{1}),...,
(\bm{s}'_{T'}, \bm{a}'_{T'}, r'_{T'})\}$ to
enhance the offline dataset. We put the augmented trajectories together to obtain the augmented dataset $\mathcal{D}_{\text{aug}}$.

\subsection{Model Implementation and Training} \label{sec:train_learn}
% We apply the trajectories in offline dataset to train the
The models used for stitching in Section \ref{sec:diffstitch} are generative model $\mathcal{G}_{\theta}$, inverse dynamic model $f_{\psi}$, reward model $f_{\phi}$ and dynamic model $f_{\omega}$, we will discuss their implementation and training in this section. 

The generative model $\mathcal{G}_{\theta}$ is designed to generate the transition states between to-stitch trajectories, and we implemented it with a diffusion model in this paper. 
Therefore, for
% a matrix $\bm{M}_s$ which represents a 
noise-masked state sequence $\tau_{s,m}$, $\mathcal{G}_{\theta}$ reconstruct it with $K$ denoising steps :
\begin{equation} \label{eq:denoise}
\begin{aligned}
\resizebox{0.5\textwidth}{!}{$
    \tau_{s,m}^{k-1} = \frac{1}{\sqrt{{\alpha}_k}} \left(\tau_{s,m}^{k} - \frac{1 - {\alpha}_k}{\sqrt{(1 - \bar{{\alpha}_k})}} \epsilon_{\theta}(\tau_{s,m}^{k}, \tau_{s,m}, k)\right) + \sqrt{1 - {\alpha}_k} \epsilon,
$}\\
\resizebox{0.2\textwidth}{!}{$\epsilon \sim \mathcal{N}(0, \textbf{I}),\text{for} \ k = K, .., 1
$},
\end{aligned}
\end{equation}

where $\alpha_i$ is the $cosine$ function of $k$ and $\bar{\alpha}_i = \prod_1^i \alpha_t$. $\mathcal{N}$ denotes the Gaussian distribution. 
$\epsilon_{\theta}$ is a neural network, we use  U-Net~\citep{ronneberger2015u} to implement it. $\tau_{s,m}^0$ is the reconstructed states without any noise. 
% Therefore, 
From Eq.~\ref{eq:denoise}, we can infer that
training $\mathcal{G}_{\theta}$ is essentially identifying the parameters of $\epsilon_{\theta}$.

To identify the parameters, we 
% Like the previous work \cite{ajay2022conditional}, we train $\mathcal{G}_{\theta}$ with the adding-noise process. Since the generative model $\mathcal{G}_{\theta}$ is diffusion, and it is designed to generate the transition states between to-stitch trajectories, we
apply the state sequence of trajectories in offline dataset, and train $\epsilon_{\theta}$ in a adding-noise procedure like previous works\cite{ajay2022conditional, janner2022planning}.
% First, we take out the state sequence of trajectories and randomly mask them.

Specifically, for an arbitrary trajectory, we first take out the state sequence of trajectories with a horizon of size $H$, \textit{i.e.}, $\tau_{s}=(\textbf{s}_{1}, \textbf{s}_{2}, ..., \textbf{s}_{H})$, and mask it with $\tau_{s,m} =
(\textbf{s}_{i},\texttt{[MASK]}, ... , \texttt{[MASK]}, \textbf{s}_{k}, \textbf{s}_{k+1}.., \texttt{[MASK]})$,
% \begin{equation}
%   \label{eq4-2}
% % \hat{\bm{M}}
% \tau_{s,m} =
% (\textbf{s}_{i},\texttt{[MASK]}, ... , \texttt{[MASK]}, \textbf{s}_{k}, \textbf{s}_{k+1}.., \texttt{[MASK]}),
% \end{equation} 
% where $\texttt{[MASK]}$ denote a masked element. T
where the mask pattern randomly conceals two intervals. We strictly ensure that the initial state of $\tau_{s}$ is unmasked, while the terminal state of $\tau_{s}$ is masked.

Next, we randomly sample a noise $\epsilon \sim \mathcal{N}(0, \mathcal{I})$ and a diffusion timestep $k \sim \mathcal{U}\{1,...,K\}$, and train $\mathcal{G}_{\theta}$ with:
\begin{equation} \label{eq: diff_loss}
    \mathcal{L}_\theta=\mathbb{E}_{k,\tau_{s} \in \mathcal{D}}{||\epsilon - \epsilon_{\theta}(\sqrt{\bar{\alpha}} \tau_s + \sqrt{1-\bar{\alpha}} \epsilon, \tau_{s,m},k) ||^2} .
\end{equation}

% \begin{equation} \label{eq: diff_loss}
%     \mathcal{L}_\theta=\mathbb{E}_{k,\tau_{s} \in \mathcal{D}}{||\epsilon - \epsilon_{\theta}(\sqrt{\bar{\alpha}} \tau_s + \sqrt{1-\bar{\alpha}} \epsilon, \tau_{s,m},k, R(\tau_{s,m})) ||^2} .
% \end{equation}

The inverse dynamic model $f_{\psi}$, reward model $f_{\phi}$, and dynamic model $f_{\omega}$ are all implemented with MLPs (Multiple Layer Perceptrons). We use states, rewards, and actions in the
trajectories of offline datasets to train them. That is, they are trained by:
\begin{equation} \label{eq:other}
    \begin{aligned}  \mathcal{L}_{\phi}&=\mathbb{E}_{(\bm{s}_{t},\bm{a}_{t},r_{t},\bm{s}_{t+1}) \sim \mathcal{D}} {\bigl [ \bm{a}_{t} - f_{\psi}(\bm{s}_{t},\bm{s}_{t+1})\bigl]}, \\
\mathcal{L}_{\psi}&=\mathbb{E}_{(\bm{s}_{t},\bm{a}_{t},r_{t},\bm{s}_{t+1}) \sim \mathcal{D}} {\bigl [ r_{t} - f_{\phi}(\bm{s}_{t},\bm{a}_{t})\bigl]}, \\    \mathcal{L}_{\omega}&=\mathbb{E}_{(\bm{s}_{t},\bm{a}_{t},r_{t},\bm{s}_{t+1}) \sim \mathcal{D}} {\bigl [ \bm{s}_{t+1} - f_{\omega}(\bm{s}_{t}, \bm{a}_{t})\bigl]}, \\
    \end{aligned}
\end{equation}

We summarize the detailed pseudo code for DiffStitch in Appendix \ref{pseudocode-section}. In training offline RL algorithms, we sample data from $\mathcal{D}^*=\mathcal{D} \cup \mathcal{D}_{\text{aug}}$ with a fixed ratio $r = (\text{number of original data}:{\text{number of augmented data}})$ between the original data from $\mathcal{D}$ and the augmented data from $\mathcal{D}_\text{aug}$ in each training batch.

\renewcommand\arraystretch{1.2}
\begin{table*}[ht]
    \vspace{5pt}

    % \caption{Offline Reinforcement Learning Performance. The performance of baseline methods are reported using the same results reported in their own papers.  Optimal and sub-optimal results of each setting are marked as \textbf{bold} and \underline{underline}, respectively.}
    \caption{
    The average normalized score of different methods. Here $\pm$ denoting the variance. The mean and standard deviation are computed over 3 random seeds. The best and the second-best results of each setting are marked as \textbf{bold} and \underline{underline}, respectively.
    }
    \vspace{8pt}
    \centering
    % \vspace{-5pt}
    % \scriptsize
    % \begin{lrbox}{\tablebox}
    % \hspace{-5pt}
    % \begin{tabular}{p{1.8cm}|r|c|l|l|l|l|l|l|l|l}
    \resizebox{0.97\linewidth}{!}
    {
        \begin{tabular}{l l ||c c c |c|| c c c |c || c |c}
            \hline
            \multirow{2}*{\textbf{Dataset}} & \multirow{2}*{\textbf{Environment}}&\multicolumn{4}{c||}{\textbf{IQL}} & \multicolumn{4}{c||}{\textbf{TD3+BC}} & \multicolumn{2}{c}{\textbf{DT}} \\
            \cline{3-12}
             % \textbf{Dataset} & \textbf{Environment}    & \textbf{Original}      & \textbf{SYNTHER}  & \textbf{TATU}   & \textbf{DStitch}  & \textbf{TD3}      & \textbf{TD3+SYNTHER}  & \textbf{TD3 + TATU}      & \textbf{TD3+DStitch}  & \textbf{DT}  & \textbf{DT++SYNTHER}   & \textbf{DT+TATU}   & \textbf{DT+DStitch} \\
            &     & Original      & SER  & TATU   & \textbf{DStitch}  & Original      & SER  & TATU   & \textbf{DStitch}  & Original     & \textbf{DStitch}  \\
             % Original
            \hline
            % \multirow{3}*{Med-Expert} 
           Med-Expert & HalfCheetah  &92.7$\pm$2.5&88.9$\pm$2.1&\textbf{94.4}$\pm$0.6&\textbf{94.4}$\pm$1.4&\underline{94.3}$\pm$0.9&86.5$\pm$8.8&89.3$\pm$3.9&\textbf{96}$\pm$0.5&\underline{90.8}$\pm$1.4&\textbf{92.6}$\pm$0.1    \\
           Med-Expert  & Hopper   &98.7$\pm$10&\underline{110.4}$\pm$1.6&93.4$\pm$17.8&\textbf{110.9}$\pm$2.9&94.8$\pm$13&\underline{104}$\pm$10.8&99$\pm$14.9&\textbf{107.1}$\pm$7&\textbf{109.9}$\pm$1.9&\underline{109.4}$\pm$1.9\\
            Med-Expert & Walker2d                       &\textbf{112.4}$\pm$0.8&\underline{111.7}$\pm$1.1&110.7$\pm$0.6&111.6$\pm$0.1&109.9$\pm$0.8&\underline{110.5}$\pm$0.3&\textbf{110.7}$\pm$0.7&110.2$\pm$0.3&\underline{108.1}$\pm$0.3&\textbf{108.6}$\pm$0.4\\
            \hline
           % \multirow{3}*{Medium} 
           Medium & HalfCheetah                       &48.5$\pm$0.3&\underline{49.3}$\pm$0.1&48.2$\pm$0.1&\textbf{49.4}$\pm$0.14&\underline{48.4}$\pm$0.1&\underline{48.4}$\pm$0.4&48.1$\pm$0.2&\textbf{50.4}$\pm$0.5&\underline{40.4}$\pm$2.2&\textbf{44.2}$\pm$0.3\\
            Medium & Hopper   &65.9$\pm$5.7&\underline{66.6}$\pm$2.4&60.3$\pm$3.6&\textbf{71}$\pm$4.2&58$\pm$2.8&56.4$\pm$1.3&
            \underline{58.3}$\pm$4.8&\textbf{60.3}$\pm$4.9&\textbf{61.5}$\pm$2.2&\underline{60.5}$\pm$4.3\\
            Medium & Walker2d                          &81.1$\pm$1.8&\textbf{85.9}$\pm$1.6&76.6$\pm$10.7&\underline{83.2}$\pm$2.2&81.4$\pm$2.3&\textbf{84.9}$\pm$0.3&75.8$\pm$3.5&\underline{83.4}$\pm$1.7&\underline{70.5}$\pm$1.6&\textbf{72}$\pm$4.9\\
            \hline
            % \multirow{3}*{Med-Replay}
            Med-Replay& HalfCheetah                    &44.1$\pm$0.5&\textbf{46.6}$\pm$0.1&44.2$\pm$0.1&\underline{44.7}$\pm$0.1&44.5$\pm$0.2&\textbf{45.2}$\pm$0.1&44.5$\pm$0.3&\underline{44.7}$\pm$0.3&\underline{39.8}$\pm$1.6&\textbf{41}$\pm$0.4\\
             Med-Replay& Hopper                        &91.4$\pm$8.1&\textbf{102.4}$\pm$0.5&79.6$\pm$7.6&\underline{102.1}$\pm$0.4&\underline{64.8}$\pm$19.2&56.8$\pm$9.9&64.1$\pm$10.5&\textbf{79.6}$\pm$13.5&\underline{83.6}$\pm$3.9&\textbf{96.1}$\pm$2\\
             Med-Replay& Walker2d                       &80.7$\pm$7&\underline{85.7}$\pm$3.6&75$\pm$12.1&\textbf{86.6}$\pm$2.8&82.4$\pm$5&\underline{89.1}$\pm$0.5&62.1$\pm$10.4&\textbf{89.7}$\pm$4.2&\underline{53.3}$\pm$11.2&\textbf{60.2}$\pm$1.8\\
            \hline
            \multicolumn{2}{c||}{Locomotion Average} &79.5&\underline{83.1}&75.8&\textbf{83.8}&75.4&\underline{75.8}&72.4&\textbf{80.2}&73.1&\textbf{76.1} \\
            \hline
            human  & pen  &79.1$\pm$28.5&\underline{88.9}$\pm$22.6&\textbf{96.8}$\pm$8.6&87.4$\pm$8.6&-3.3$\pm$0.4&-2.8$\pm$1.5&\underline{7.3}$\pm$14.1&\textbf{29.8}$\pm$6.9&\underline{62.8}$\pm$2.1&\textbf{70}$\pm$6.8 \\
            cloned  & pen  &45.8$\pm$29.9&\underline{52.5}$\pm$27.9&45.3$\pm$23.4&\textbf{64}$\pm$29.6&-3.1$\pm$0.4&\underline{-2.8}$\pm$0.1&-3$\pm$0.3&\textbf{11.3}$\pm$6.2&\underline{57.7}$\pm$3.3&\textbf{59.7}$\pm$9.6\\
            %Large   & Maze2d   &&&&&&&&&&&\\
            \hline
            \multicolumn{2}{c||}{Pen Average} &62.5&70.7&\underline{71.1}&\textbf{75.7}&-3.2&-2.8&\underline{2.2}&\textbf{20.6}&60.3&\textbf{64.9} \\
            \hline

            % Med-Replay& Hopper 
            % Med-Replay& Hopper 
            % \multicolumn{2}{c|}{Kitchen-Mixed-V0} 
            % Mixed-V0 & Kitchen &&&&&&&&&&& \\
            % % \hline
            % % \multicolumn{2}{c|}{Kitchen-partial-v0} 
            % partial-v0& Kitchen &&&&&&&&&&& \\
            % \hline
            % \multicolumn{2}{c|}{Kitchen Average} &&&&&&&&&&& \\
            \hline
        \end{tabular}
    }
    \label{tab:main_result}
\end{table*}

\section{Experiments}
In this section, we will discuss the experimental settings and the performance of DiffStitch. 

% We conduct extensive experiments to evaluate the DiffStitch. Specifically, we evaluated the performance of DiffStitch across different offline RL algorithms and environments, conducted an in-depth analysis of its effectiveness, and examined the factors influencing its performance.

% In this section, we enhance various offline reinforcement learning (RL) algorithms with DiffStitch and conduct extensive experiments on the D4RL MuJoCo and Kitchen benchmarks.
% \longting{the following sentences are unnecessary}
% Sections 5.1 and 5.2 present the augmentation of various offline RL algorithms to demonstrate the effectiveness of stitching transitions generated by DiffStitch. Sections 5.3 and 5.4 include a detailed ablation study aimed at answering the following questions: (1) How does DiffStitch compare with single-way generation? (2) Is estimating the stitching timestep necessary? (3) How does different trajectory selection strategies affect performance? In Section 5.5, we compare DiffStitch with other data augmentation methods to highlight the superiority of our proposed algorithm.

% \longting{section 1. give the dataset or environment; section 2: implementation details are necessary, e.g., hyperparameters}

% \longting{cover baselines and validation algorithm}
\subsection{Experiment Settings}
% \subsection{Evaluation Environments} 
\paragraph{Evaluation Environments}
We evaluate DiffStitch on a wide range of domains in the D4RL benchmark \cite{fu2020d4rl}, including MuJoCo tasks and Adroit tasks. Specifically, we evaluate the performance of DiffStitch on three environments in MuJoCo tasks: HalfCheetah, Hopper, and Walker2d. Each environment includes three types of datasets: Medium-expert, Medium, and Medium-replay, in which the Medium-expert is composed of both expert demonstrations and suboptimal data, Medium-replay and Medium are collected when an unconverged policy \cite{haarnoja2018soft} interacted with the environment. We evaluate the performance of DiffStitch with the dataset of Human and Cloned in Adroit-pen tasks, which aims to control a robot to twirl a pen. The Human dataset is composed of human demonstrations and the Cloned dataset is collected by an imitation policy on the demonstrations. Compared to MuJoCo tasks, Adroit-pen has a larger observation space (45 dimensions) and action space (24 dimensions), making it more challenging.

% We compare the performance of offline RL algorithms with and without transitions generated by DiffStitch. Additionially, we conduct a  comparative analysis to evaluate the effectiveness of DiffStitch in comparison to other generative data augmentation baselines.

% \textbf{Gym-MuJoCo.} The Mujoco tasks include three different environments (Walker2d, halfcheetah, hopper), each including three types of datasets(medium-expert, medium-replay, medium). The “medium-expert” dataset is generated by mixing equal amounts of expert demonstrations and suboptimal data. The “medium-replay” dataset consists of recording all samples in the replay buffer observed during training until the policy reaches the “medium” level of performance. The "medium" dataset is generated by first training a policy online using Soft Actor-Critic\cite{haarnoja2018soft}, early-stopping the training, and collecting 1M samples from this partially-trained policy. 

% \textbf{Adroit-pen.} The Adroit-pen task involves controlling a 24-DoF simulated Shadow Hand robot tasked with twirling a pen. In this environment, we evaluate it on two kinds of datasets: "human" and "cloned". The "human" dataset includes a small amount of demonstration data from a human. The "cloned" dataset is generated by training an imitation policy on
% the demonstrations, running the policy, and mixing data at a 50-50 ratio with the demonstrations. Compared to MuJoCo tasks, Adroit-pen has a larger observations space(45 dimensions) and action space(24 dimensions), making it more challenging. 

% \subsection{Data Augmentation Baselines} 
\paragraph{Baselines}
Since we aim to augment the offline dataset to improve the performance of offline RL algorithms,  
we compare our method with two recent data augmentation algorithms in offline RL, SER\cite{lu2023synthetic} and TATU\cite{zhang2023uncertainty}, which have demonstrated outstanding performance. SER is designed to utilize the diffusion model for learning the distribution of transitions in the offline dataset and generating new transitions. However, since these transitions are generated without any conditions, SER cannot generate complete trajectories. On the other hand, TATU employs a forward dynamic model to simulate transitions based on the offline dataset. Nevertheless, TATU is specifically designed to roll out forward transitions in very  limited steps to ensure that the generated transitions remain within the distribution, thereby making it challenging to generate long trajectories.

% TATU uses a forward dynamic model to rollout transitions and incorporates mechanisms to guarantee that the generated transitions do not fall in out-of-distribution(OOD) regions. SER trains a diffusion model $\mathcal{P_{\varsigma }(D)}$ to model the distribution of each transition tuple $(s,a,r,s')$ and resample from $\mathcal{P_{\varsigma }(D)}$ to generate more transition tuples.

% It is important to note, as depicted in Figure \ref{fig:method_difference}, that the data generated by SER and TATU are not able to augment sequential decision-making methods such as Decision Transformer (DT). DT's input is in the form of $\{(R_i, s_i, a_i),..., (R_{i+H},s_{i+H},a_{i+H})\}$ where $R_{i}$ denotes return-to-go of timestep $i$ in the trajectory. TATU only rolls out forward transitions for less than five steps, i.e., $s, \widetilde{s_{1}},...,\widetilde{s_{5}} $, starting from $s \in \mathcal{D}$. If we employ the data augmented by TATU to train Decision Transformer, then the return-to-go must be calculated from $\widetilde{s_{5}}$ with a sum of zero. However, this does not accurately represent the actual return-to-go starting from $\widetilde{s_{5}}$, as there may still be hundreds of steps beyond $\widetilde{s_{5}}$. SER does not generate transitions conditioned on any state from the original dataset $\mathcal{D}$; it generates transition tuples with a length of 1 via a diffusion model, which also cannot effectively augment Decision Transformer (Decision Transformer's default input horizon length is 20). 

% \subsection{Evaluation Algorithms} 
\paragraph{Evaluation Algorithms} 
To assess the quality of the augmented data, we utilize the augmented dataset to train three groups of offline RL algorithms: TD3+BC \cite{fujimoto2021minimalist}, IQL \cite{kostrikov2021offline}, and DT \cite{chen2021decision}. All of these methods have been widely employed in previous studies due to their popularity \cite{prudencio2023survey}. Here IQL is a one-step method, which first learns the value function and then derives the policy based on the learned value function. TD3+BC is a imitation learning method, which jointly learns the policy and value function. DT is a trajectory optimization method, which takes trajectory data as input and applies Transformer \cite{vaswani2017attention} to model the distribution of trajectories in the offline dataset.

\paragraph{Implementaion Details}
For the generative model, the horizon of generation is set to 100 for MuJoCo tasks, and 55 for Adroit-pen tasks. 
% Following \cite{ajay2022conditional}, we implement $u_{\theta}$ of the generative model via U-Net~\citep{ronneberger2015u} to reconstruct the states from Gaussian noise. 
The denoising step $K$ is set to 100. We implement the inverse dynamics model $f_{\psi}$ with a 2-layer MLP, and the dynamic model $f_{\omega}$, reward model $f_{\phi}$ with a 4-layer MLP. For the stitching, we set the qualification threshold used for data selection $\delta$ ranging from $[1, 16]$ depending on the dataset.
% as the denoise network $\psi_\theta(\cdot,\cdot)$ and the return predictor $\mathcal{J}_\phi(\cdot, \cdot)$, and adopt a linear layer with $Sigmoid$ as the activation function as the projector $f(\cdot)$.
 Our model is trained on a device with A5000 GPUs(24GB GPU memory, 27.8 TFLOPS computing capabilities, AMD EPYC 7371 16-Core Processor, optimized by Adam~\citep{kingma2014adam} optimizer. 
 % Details of hyper-parameters are shown in Appendix \ref{hyperparameter-sec}
 
 %\longting{\cref{appendix:hyper}}.

% \subsection{Performance on D4RL Datasets}
\vspace{-10pt}
\subsection{Main Results}
\label{sec:main-result}
% We leverage three offline RL algorithms to evaluate our DiffStitch. Specifically, 
To evaluate the performance of DiffStitch and compare its performance with the baselines,
we train TD3+BC and IQL with the augmented data generated by DiffStitch, SER, and TATU, as well as the original offline data. This allows us to compare the effectiveness of the augmented data.  Additionally, we use the data generated by DiffStitch and the original offline data to train DT in order to assess the effectiveness of DiffStitch. We did not employ the data generated by SER and TATU to train DT because either they were unable to generate trajectories or could only produce very short trajectories, which makes the data generated by them inapplicable to augment decision transformer(DT). The results are presented in Table \ref{tab:main_result}. Each result is obtained by conducting 10 trials with different seeds, and reported with the normalized average returns ~\citep{fu2020d4rl} and standard deviation.

\begin{figure}[t]
    % \vskip 0.2in
    \begin{center}
    \vspace{5pt}\centerline{\includegraphics[width=1\columnwidth]{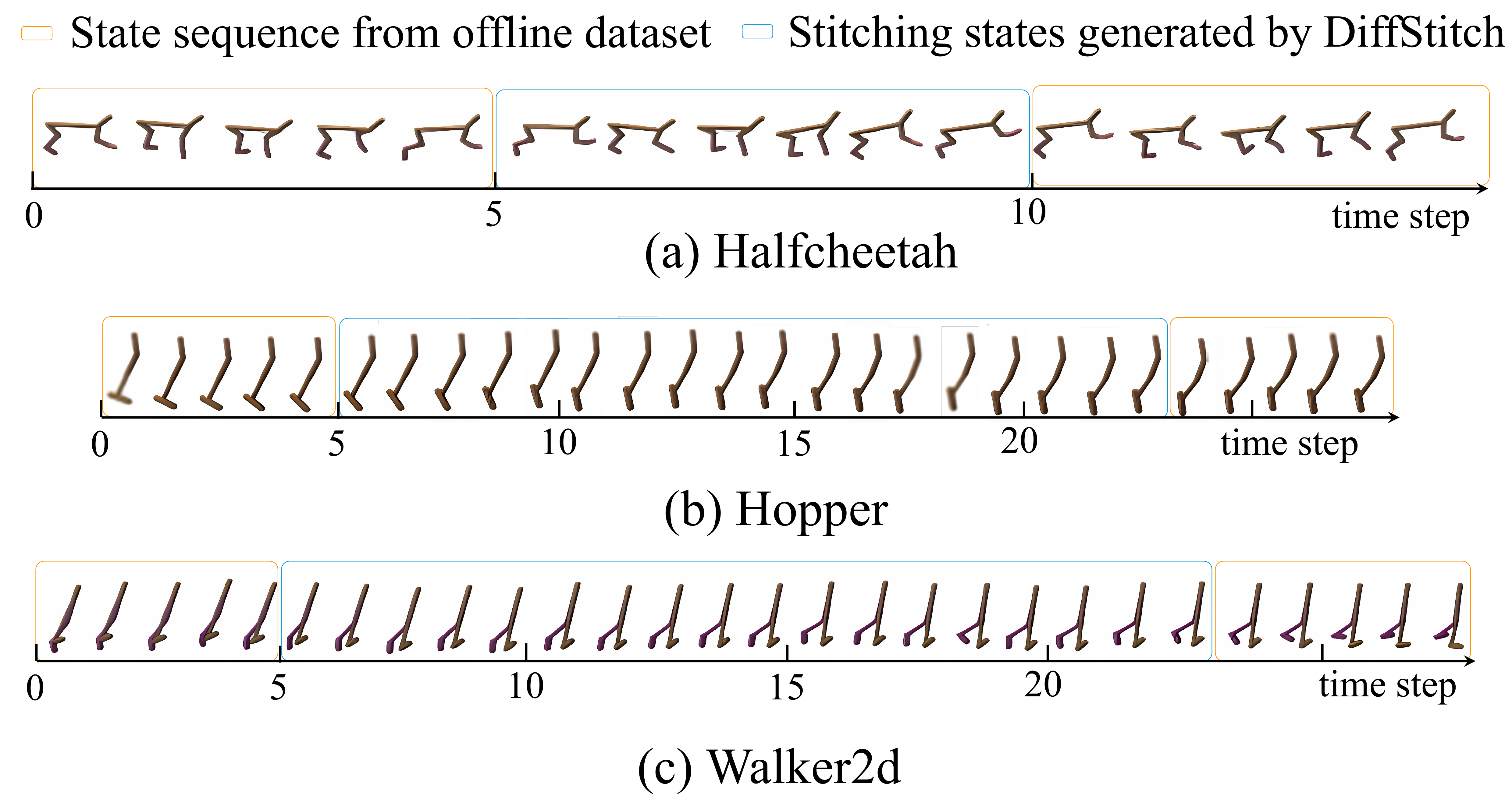}}
     \vspace{-5pt}
      \caption{The stitching trajectory.}
      \label{fig:vis}
      \vspace{-25pt}
    \end{center}
   
\end{figure}

\begin{figure}[t]
    % \vskip 0.2in
    \begin{center}
    % \centerline{\includegraphics[width=\columnwidth]{figs/contrast_RdYlBu.pdf}}
    \centerline{\includegraphics[width=\columnwidth]{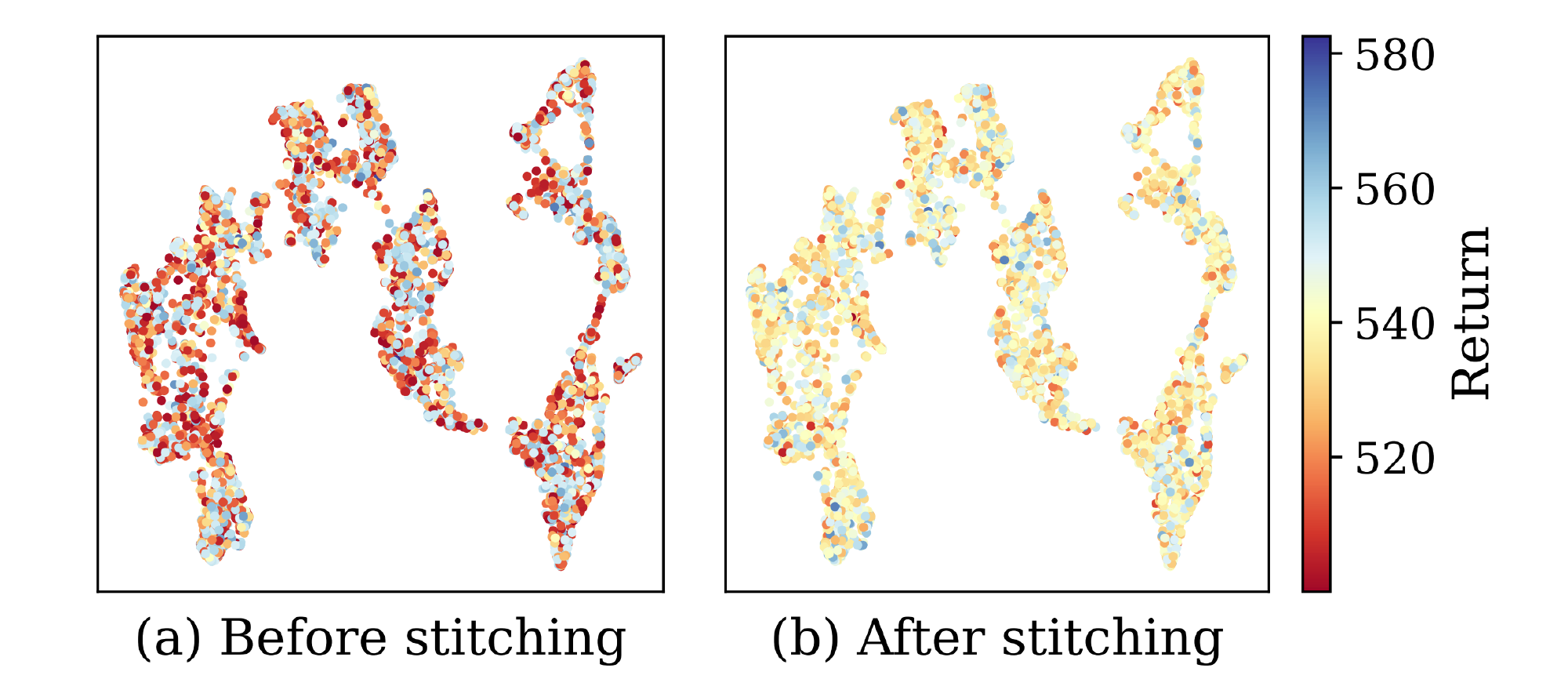}}
    \vspace{-10pt}
      \caption{Return of states before and after stitching}
      \label{fig:return}
      \vspace{-20pt}
    \end{center}
\end{figure}

From Table \ref{tab:main_result}, we can observe: (1) Our DiffStitch achieves the best performance in most cases, and obtains the best average performance in both MuJoCo tasks and Adroit tasks. This demonstrates the data generated by our method has higher quality than the data generated by SER and TATU in most cases; (2) Our DiffStitch demonstrates superior performance on challenging tasks. As Table \ref{tab:main_result} illustrated, It is easier for DiffStitch to achieve significant improvement in the cases where the original score is relatively low, like the cases on pen, the case of TD3+BC on Med-Replay Hopper. The low score in these cases demonstrates that offline RL algorithms have difficulty exploring high-reward regions. With the trajectory stitching, our DiffStitch potentially offers high-reward training samples for them. (3) Both our DiffStitch and SER utilize the diffusion model for data augmentation. However, our DiffStitch achieves better performance than SER in 14 out of 22 cases, indicating that augmenting data from the perspective of trajectory stitching is more effective in most scenarios.(4)Offline RL methods trained on augmented data have better performance than the one trained on original data, which demonstrates enhancing the performance of offline RL via data augmentation is effective.

\subsection{Further Analysis}
As observed in Section \ref{sec:main-result}, DiffStitch outperforms baseline augmentation methods in most cases. We assume two features mainly contribute to the advancements: (1) The generated stitching sub-trajectories are consistent with the environment dynamics, and (2) DiffStitch successfully transforms low-reward trajectories to high-reward trajectories. Both features ensure the provision of high-quality data for offline RL algorithms to learn from, resulting in improved performance.
% As it is demonstrated in Table \ref{tab:main_result}, DiffStitch is effective in most cases, which improves the performance of different offline RL algorithms by the augmented data. We assume there are two reasons for its achievement: The stitching naturally resembles the transitions found in offline datasets, smoothly connecting sub-trajectories with low rewards to those with high rewards. This provides valuable samples for offline RL algorithms to learn how to transition effectively from states with low rewards to states with high rewards. 
% To validate the quality of stitching, 
To investigate whether the stitching sub-trajectories conform to environment dynamics, we randomly select three augmented trajectories, one for each of HalfCheetah, Walker2d, and Hopper. We visualize the robot states at each time step, and the results are presented in Figure \ref{fig:vis}. We can observe the stitching states are quite natural and roughly align with the locomotion patterns of these robots.
To check whether the generated samples transform low-reward trajectories to high-reward trajectories, we randomly take the states from the left side of stitching trajectories and compare their returns before and after stitching in Figure \ref{fig:return}. We can observe that the returns of the majority of states have improved after stitching. 
% As a result, for those trajectories starting from those states,
As the returns are the accumulation of discounted future rewards, the improvement in return indicates that the originally lower-reward trajectories starting from those states are transformed into higher-reward trajectories, serving as more valuable data in policy learning.

% groups of samples in \longting{Medium Hopper?}, in which a robotic-legged creature learns to jump from the offline dataset. 
% we illustrate the trajectories before and after stitching in Figure \ref{fig:hopper_vis}. From Figure \ref{fig:hopper_vis}, we can observe the stitching is much more natural, and it roughly aligns with the locomotion patterns of a leg. To validate changes in return, we randomly take the states that appear before the stitching sub-trajectories in the time step, and visualize their corresponding returns before and after stitching in Figure \ref{fig:return}. From Figure \ref{return}, we can observe that the returns of the majority of states have shown improvement after stitching, offering valuable insights for a policy to learn effective strategies in reaching the high-reward region.

% In Figure \ref{fig:further}, we visualize the states that were stitched to another trajectory in halfcheetah-medium-v2, color-coded according to the discounted return-to-go values(with discount reward gamma equals to 0.99). It can be shown that after augmenting the dataset with DiffStitch, the discounted return-to-go values of stitched states have increased, which makes the dataset more optimum, improving the final performance of offline RL methods. 

% In Figure \ref{fig:hopper_vis}, we render a 

\begin{figure}[t]
    % \vskip 0.2in
    \begin{center}
    \centerline{\includegraphics[width=\columnwidth]{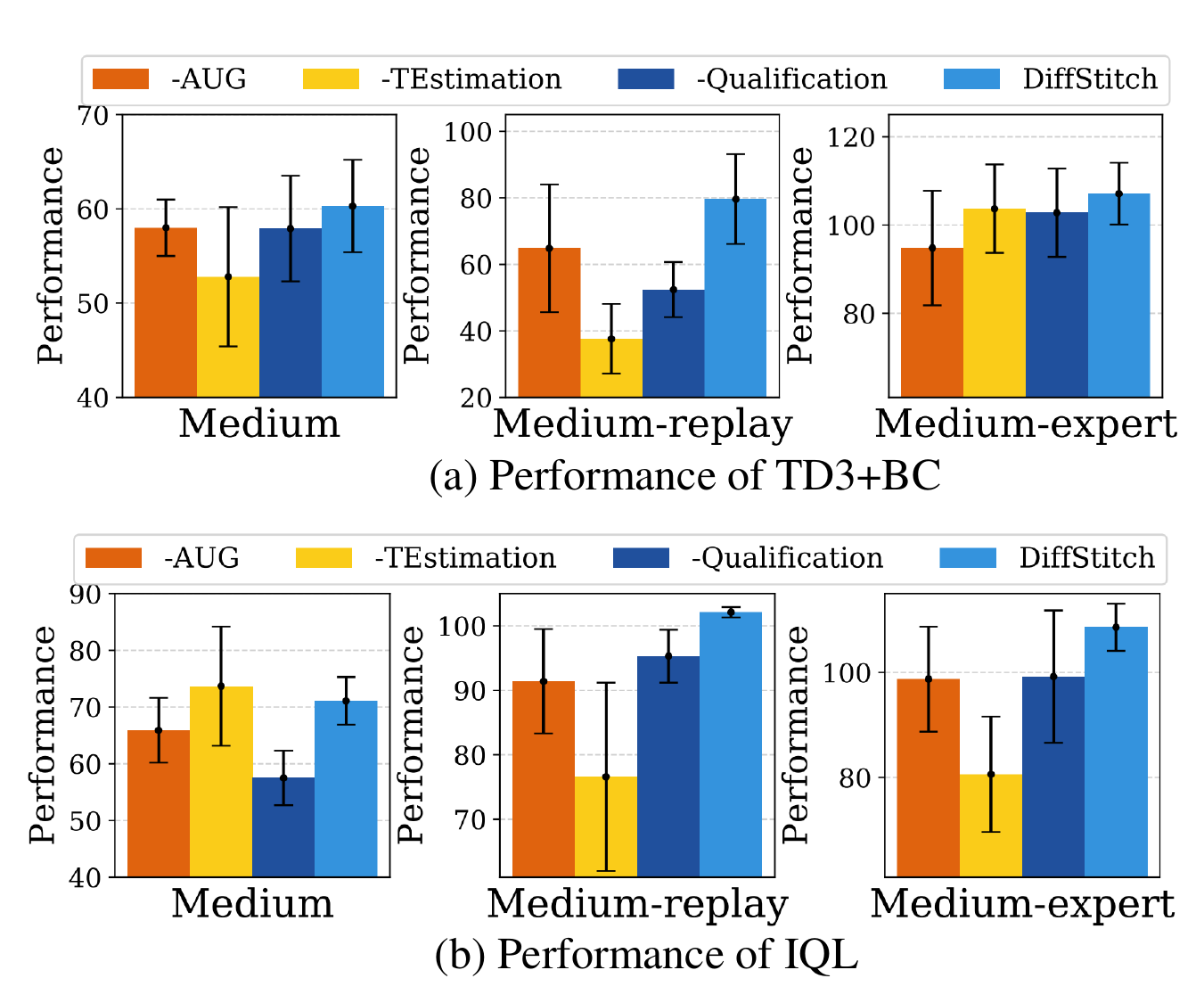}}
    \vspace{-10pt}
      \caption{Ablation study on the hopper }
      \label{fig:ablation_study}
    \end{center}
    % \vskip -0.2in
    \vspace{-15pt}
\end{figure}

\subsection{Ablation Study}  
To investigate the impact of the modules in DiffStitch, we apply IQL and TD3+BC to conduct ablation studies on the environment of Hopper. Specifically, we have the variants:
\begin{itemize}[leftmargin=10pt]
    \item \textbf{-AUG} remove the augmented data, and directly train the offline RL algorithms with the original data.
    \item \textbf{-TEstimation} remove the step estimation module from DiffStitch. In this case, the number of stitching steps is set to a constant value 40. 
    \item \textbf{-Qualification} remove the qualification module from DiffStitch. In this case, all the data generated by DiffStitch are directly used to train the offline RL algorithms.
\end{itemize}

The results are illustrated in Figure \ref{fig:ablation_study}. We can observe from Figure \ref{fig:ablation_study}: (1) the performance of DiffStitch is better than -TEstimation in most cases. This demonstrates that estimating the number of stitching steps is important. Conversely, utilizing a fixed number of states to stitch trajectories might be detrimental, leading to a degradation in the original performance; (2) the performance of DiffStitch is better than -Qualification in all cases. That demonstrates the crucial role of data selection;
% We also conducted a comprehensive investigation into the influence of the qualification threshold, as outlined in Appendix \ref{quali-hyperparame}   
(3) DiffStitch outperforms -AUG in all the cases, which highlights the benefits of data augmentation in offline RL tasks.

\subsection{Parameter Study} 
% \subsection{Other factors}
To further investigate the factors that influence the performance of DiffStitch, we investigate the impact of data ratio and qualification threshold $\delta$.

% \textbf{Original data ratio $r$.} 
% \paragraph
% \subsection{The Impact of Data Ratio}
\paragraph{The impact of data ratio $r$}
DiffStitch enhances the effectiveness of offline RL algorithms by generating augmented data. In the training of offline RL algorithms, both the augmented dataset and original offline dataset are used. Consequently, the ratio of the original data to the augmented data might impact the performance of DiffStitch. To investigate that, we conduct experiments on Halfcheetah-Medium-expert with different ratios. Specifically, we set the ratio to $\{0:1, 1:1, 2:1, 4:1, 9:1, 1: 0\}$, and other settings remain the same to test the performance. The results are illustrated in Table \ref{tab:ratio-table}. Except for the effectiveness of the augmentation, we can also observe in Table \ref{tab:ratio-table} that both an excessively large ratio and an excessively small ratio have detrimental effects on the performance of DiffStitch. We assume the reasons are as follows: When ratio is excessively large(1:0), the benefit brought by augmentation is abandoned. When data ratio is excessively small(0:1), too much noise are introduced and agents cannot effectively learn a policy. 

% because of the benefit brought by augmentation is abandoned in the case of the excessively large ratio (1:0), but much noise are introduced in the case of the excessively small ratio (0:1). 

% The original data ratio $r$ controls the amount of data sampled from the original dataset $\mathcal{D}$ in every mini-batch during training, influencing the final performance of offline RL methods. To analyze how $r$ affects the performance of offline RL methods, we reran IQL and TD3\_BC on halfcheetah-medium-expert-v2 with different ratios $r\in {1:1, 2:1, 4:1, 9:1}$. The results are summarized in Table\ref{tab:ratio-table}. The optimal ratio $r$ varies across different datasets and offline rl algorithms. 

% \renewcommand\arraystretch{1}
\begin{table}[t]
  \centering
  % \caption{Results of TD3+BC, IQL and DT trained on DiffStitch.}
  \caption{The impact of data ratio $r$.}
  \vspace{10pt}
  \resizebox{0.28\textwidth}{!}{
  \begin{tabular}{c| c c c}
    \hline
    ratio & TD3+BC & IQL & DT \\
    \hline
    0:1 & 0.1$\pm$0.9 & -0.5$\pm$1.0& 1.2$\pm$0.3 \\
    1:2 & 91.7$\pm$2.3 & 91.5$\pm$3.7 & \textbf{92.6}$\pm$0.1 \\
    1:1 & 94.1$\pm$1.2 & 91.7$\pm$3.7 & 92.5$\pm$0.6\\
    2:1 & 93.2$\pm$2.8& 93.9$\pm$1.0&91.5$\pm$1.0 \\
    4:1 & \textbf{96}$\pm$0.5 & \textbf{94.4}$\pm$1.4 &  92.4$\pm$0.5\\
    1:0 & 94.3$\pm$0.9&92.7$\pm$2.5& 90.8$\pm$1.4\\
    \hline
  \end{tabular}
  }
  \vspace{-10pt}
  \label{tab:ratio-table}
\end{table}

\paragraph{The impact of qualification threshold $\delta$.}

\label{quali-hyperparame}
The qualification threshold $\delta$ determines the decision to exclude generated transitions that deviate from environmental dynamics, guaranteeing that the generated transitions remain within in-distribution (ID) regions. To investigate the impact of $\delta$, we conduct experiments on Halfcheetah-Medium-Expert, and test the performance of our method under TD3+BC, IQL and DT. The experiment results are presented in Table \ref{tab:quali-table}. We can observe from Table \ref{tab:quali-table} that a too small $\delta$ or a too large $\delta$ decrease the performance of all the evaluation algorithms (TD3+BC, IQL and DT). We assume that a smaller $\delta$ provides stronger assurance of adherence to the distribution but may be overly conservative, whereas a larger $\delta$ may permit transitions that deviate from environmental dynamics, falling into OOD regions. 

% e perform a case study on the MuJoCo dataset halfcheetah-medium-expert-v2 to investigate the impact of $\delta$ on the final performance of TD3+BC, IQL and DT. The results are represented in Table \ref{tab:qualification-table}. It can be observed that the optimal qualification threshold for the three algorithms is 2 and 3. This suggests that the selection threshold should be neither too large nor too small.

\begin{table}[t]
  \centering
  \caption{The impact of qualification threshold $\delta$.}
  \vspace{10pt}
  \resizebox{0.28\textwidth}{!}{
  \begin{tabular}{c| c c c}
    \hline
    $\delta$ & TD3+BC & IQL & DT \\
    \hline
    1.0 & 93.8$\pm$2.0 & 92$\pm$1.7 &93.1$\pm$0.2  \\
    2.0 &\textbf{96}$\pm$0.5&94.4$\pm$1.4&92.6$\pm$0.1 \\
    3.0 &91.7$\pm$3.2&\textbf{95}$\pm$0.5&\textbf{92.7}$\pm$0.6 \\
    4.0 & 94.4$\pm$0.8 & 91.1$\pm$2.7 & 92.6$\pm$ 1 \\
    % \hline
    \hline
  \end{tabular}
  }
  \vspace{-15pt}
  \label{tab:quali-table}
\end{table}

% \renewcommand\arraystretch{0.5}
% \begin{table}[h]
%   \centering
%   \caption{Results of TD3+BC,IQL and DT trained on DiffStitch }
%   \vspace{10pt}
%   \resizebox{0.5\textwidth}{!}{
%   \begin{tabular}{c| c c c}
%     \hline
%     $r$ & TD3+BC & IQL & DT \\
%     \hline
%     1.0 &93.8$\pm$2.0&92$\pm$1.7&93.1$\pm$0.2& \\
%     2.0 &\textbf{96}$\pm$0.5&94.4$\pm$1.4&92.6$\pm$0.1& \\
%     3.0 &91.7$\pm$3.2&\textbf{95}$\pm$0.5&\textbf{92.7}$\pm$0.6& \\
%     4.0 &94.4$\pm$0.8&91.1$\pm$2.7&92.6$\pm$1& \\
%     \hline
%   \end{tabular}
%   }
%   \vspace{-15pt}
%   \label{tab:qualification-table}
% \end{table}

\subsection{Additional Study on Small Samples} 
To investigate the performance of DiffStitch on the dataset of limited samples, we randomly select 20\% of the samples from the original dataset, and augment the dataset with DiffStitch. Then, we train IQL, TD3+BC, and DT with the augmented dataset and compare the performance with the baselines. The experiments are conducted on HalfCheetah-Medium-Expert and 
Hopper-Medium-Replay, which covers different environments and different dataset qualities.
The results are illustrated in Table \ref{tab:20percent-result}. We can observe from Table \ref{tab:20percent-result}: (1) in the easier task (HalfCheetah-Medium-Expert), DiffStitch leverages 20\% data to achieve a performance close to using 100\% of the data in Table \ref{tab:main_result}, which demonstrates the effectiveness of Diffstitch; (2) In the challenging tasks (such as Hopper-Medium-Replay), Diffstitch demonstrates a significant advantage over the baselines. This highlights the superiority of Diffstitch in tackling difficult tasks. By stitching together low-reward trajectories with high-reward trajectories, Diffstitch effectively compensates for the limitations of the dataset.

\begin{table}[t]
    \caption{ The performance on small samples. $\pm$ denoting the variance. The mean and standard deviation are computed over 3 random seeds. The best and the second-best results of each setting are marked as \textbf{bold} and \underline{underline}, respectively.
    }
    \vspace{10pt}
    \centering
    \resizebox{1.01\linewidth}{!}
    {
        \begin{tabular}{c | l l ||c c c |c} 
            \hline
            % & \multirow{2}*{\textbf{Dataset}} & \multirow{2}*{\textbf{Environment}}&\multicolumn{4}{c}{\textbf{IQL}} \\
            % \cline{4-7}  
            % Algorithm
            & Dataset& Environment& Original& SER& TATU& \textbf{DiffStitch}\\
            \hline 
            \multirow{3}* {\rotatebox{90}{IQL}} & Med-Expert & HalfCheetah  &89.3$\pm$4.3&62.5$\pm$2.5&\textbf{91.5}$\pm$2.6&\textbf{91.5}$\pm$1.5
             \\
             & Med-Replay  & Hopper   &\underline{51}$\pm$3.9&13$\pm$1.1&49.6$\pm$13.2&\textbf{67.8}$\pm$13.5
             \\
             \cline{2-7}
             &   \multicolumn{2}{c||}{ Average}  &70.2&37.8&\underline{70.6}&\textbf{79.7}
             \\
            \hline
            \multirow{3}* {\rotatebox{90}{TD3+BC}} & Med-Expert & HalfCheetah  &\underline{81.2}$\pm$8.9&65.1$\pm$7.5&80.2$\pm$6.3&\textbf{91.6}$\pm$3.4
             \\
             & Med-Replay  & Hopper   &\underline{52}$\pm$10.1&13.5$\pm$1.3&6.7$\pm$3.7&\textbf{64}$\pm$25 
             \\
             \cline{2-7}
              &   \multicolumn{2}{c||}{ Average}   &\underline{66.6}&39.3&43.5&\textbf{77.8}
             \\
            \hline
             \multirow{3}* {\rotatebox{90}{DT}} & Med-Expert & HalfCheetah  &\underline{88.7}$\pm$2.4&-&-&\textbf{92.6}$\pm$0.16
             \\
             & Med-Replay  & Hopper   &\underline{49.3}$\pm$8.3&-&-&\textbf{75.1}$\pm$11.6
             \\
             \cline{2-7}
             &   \multicolumn{2}{c||}{ Average}   &\underline{69}&-&-&\textbf{83.9}
             \\
            
            % \multicolumn{2}{c||}{Locomotion Average} &70.2&37.8&70.6&\textbf{79.7}
            % \\
            %  \hline
            % \multirow{2}*{\textbf{Dataset}} & \multirow{2}*{\textbf{Environment}}&\multicolumn{4}{c||}{\textbf{TD3+BC}} \\
            % \cline{3-6}  
            %  & & Original& SER& TATU& \textbf{DiffStitch}\\
            % \hline 
            % Med-Expert & HalfCheetah  & 81.2\pm8.9&65.1 \pm 7.5&80.2\pm 6.3&\textbf{91.6}\pm3.4        
            \hline 
        \end{tabular}
    }
    \vspace{-15pt}
    \label{tab:20percent-result}
\end{table}

% In this section, we aim to illustrate the effectiveness of the DiffStitch in augmenting small datasets. We randomly select 20\% of the data from the original dataset $\mathcal{D}$ and label it as $\mathcal{D}_{\text{20\%}}$. We then stitch trajectories using DiffStitch on $\mathcal{D}_{\text{20\%}}$ and evaluate the performance of TD3\_BC. The outcomes for both halfcheetah-medium-expert-v2 and hopper-medium-replay-v2, with and without the augmentation of DiffStitch, are presented in Table \ref{tab:20percent-result}. The results indicate a significant improvement in the performance of TD3\_BC when trained with augmented data compared to the performance trained with only 20\% of the data from $\mathcal{D}_{\text{20\%}}$.

\section{Conclusion and Future Work}
In this paper, we introduce DiffStitch, which enhances the learning of offline RL algorithms by stitching together low-reward trajectories with high-reward trajectories.  % Leveraging the "First Imagine, then Stitch" technique and a rule-based discriminator, our method accurately determines the number of states for connecting two trajectories while eliminating low-quality data. 
Diverging from conventional approaches that primarily support one-step methods and imitation learning methods, our Diffstitch extends its utility to facilitate the learning of trajectory optimization methods such as Decision Transformer(DT). Empirical evaluations conducted on the D4RL benchmarks demonstrate a significant performance boost for base offline RL algorithms.

For future work, exploring how to choose and concatenate two trajectories will be an interesting direction. While our current approach involves concatenating low-reward trajectories with high-reward trajectories, there undoubtedly exist superior concatenation strategies waiting to be explored.

\section*{Border Impact}
This paper presents work whose goal is to advance the field of Machine Learning. There are many potential societal consequences of our work, none which we feel must be specifically highlighted here.

\nocite{langley00}

\bibliography{reference}  
\bibliographystyle{icml2024}

%%%%%%%%%%%%%%%%%%%%%%%%%%%%%%%%%%%%%%%%%%%%%%%%%%%%%%%%%%%%%%%%%%%%%%%%%%%%%%%
%%%%%%%%%%%%%%%%%%%%%%%%%%%%%%%%%%%%%%%%%%%%%%%%%%%%%%%%%%%%%%%%%%%%%%%%%%%%%%%
% APPENDIX
%%%%%%%%%%%%%%%%%%%%%%%%%%%%%%%%%%%%%%%%%%%%%%%%%%%%%%%%%%%%%%%%%%%%%%%%%%%%%%%
%%%%%%%%%%%%%%%%%%%%%%%%%%%%%%%%%%%%%%%%%%%%%%%%%%%%%%%%%%%%%%%%%%%%%%%%%%%%%%%
\newpage
\appendix
\onecolumn
\section{Experimental Details}
\subsection{Pseudocode of DiffStitch}
\label{pseudocode-section}
\begin{algorithm}[h]
   \caption{DiffStitch}
   \label{alg:psudo-algo}
\begin{algorithmic}[1]
    
   \STATE {\bfseries Input:} offline dataset $\mathcal{D}$, iterations $N$, Discriminator threshold $d$
   \STATE Initialize $\mathcal{D}_{\text{aug}} \gets \emptyset$.

   \STATE Train generative model $\mathcal{G}_{\theta}$ on $\mathcal{D}$ using Eq.(\ref{eq: diff_loss})
   \STATE Train Dynamic model $f_{\omega}$, inverse dynamic model $f_{\psi}$, and reward model $f_{\phi}$ on $\mathcal{D}$ using Eq.(\ref{eq:other})
   \FOR{$i=1$ {\bfseries to} $N$}

   \STATE Sample trajectories $\tau_{s}$, $\tau'_{s}$ from $\mathcal{D}$ 
   \STATE Imagine $\tau^m_{s}=(s_{T}, \bm{s}^m_{1}, ..., \bm{s}^m_{H-1})$ starting from $\bm{s}_{T}$ 

   % \IF{$\text{U}(\{ i|\text{dist}(\bm{s}^m_{i}, \bm{s}'_{1}) \leqslant \alpha\}) = \emptyset$}
   % \STATE Continue 
   % \ENDIF

    % \IF{$\max_{i}(\{ \text{sim}(\bm{s}^m_{i}, \bm{s}'_{1}) \}) < \alpha$} 
    % \STATE Continue 
    % \ENDIF

    \STATE Set timestep $\Delta = \argmax_{i} \text{sim}(\bm{s}_{i}^m, \bm{s}'_{1})$
   % \STATE Sample timestep $\Delta \sim \text{U}(\{ i|\text{dist}(\bm{s}^m_{i}, \bm{s}'_{1}) \leqslant \alpha\})$
   \STATE Create $\tau_{s,m} =(s_{T}\underbrace{, \ldots, }_{\Delta \text{ Masks}} , s'_{1}, ...)$

   \STATE Generate state sequence $\tau_{s}^s =\mathcal{G}_{\theta}(\tau_{s,m})$

    \STATE Wrap-up states Eq.\ref{eq:action} and Eq.\ref{eq:reward}, obtain $\tau_{gen}$ 

   \STATE For each $(\bm{s}_{t}, \bm{a}_{t}, \bm{s}_{t+1})\in \tau_{gen}$, sample $\hat{\bm{s}}_{t+1}^q \sim f_{\omega}(\bm{s}_{t},\bm{a}_{t})$
   \IF{$\max\{ ||\hat{\bm{s}}_{t+1}^q - \bm{s}_{t+1}||^2 > d\}$} 
   \STATE continue 
   \ENDIF
   \STATE
   $\mathcal{D}_{\text{aug}} \gets \mathcal{D}_{\text{aug}} \cup \tau_{gen}$
   \ENDFOR
   \STATE obtain $\mathcal{D}^* \gets \mathcal{D} \cup \mathcal{D}_{\text{aug}}$
\end{algorithmic}
\end{algorithm}

\subsection{Experimental Details} 
\textbf{Validation Offline RL algorithms.} For the validation offline RL algorithms IQL\cite{kostrikov2021offline} and Decision Transformer\cite{chen2021decision}, we use the 'CORL: Research-oriented Deep Offline Reinforcement Learning Library'\cite{tarasov2022corl} codebase. For TD3+BC\cite{fujimoto2021minimalist}, we use the author-provided code from GitHub. For IQL and TD3+BC, we train them for $10^6$ gradient steps across all tasks. For Decision Transformer(DT), we train it for $5\times10^4$ gradient steps on MuJoCo-Gym tasks and $10^5$ gradient steps on Adroit-pen tasks.

\end{document}